\documentclass[journal, 10pt]{IEEEtran}
\IEEEoverridecommandlockouts

\usepackage{amsmath,amssymb,amsfonts,mathtools}
\usepackage{algorithm}
\usepackage{array}
\usepackage[caption=false,font=normalsize,labelfont=sf,textfont=sf]{subfig}
\usepackage{textcomp}
\usepackage{stfloats}
\usepackage{url}
\usepackage{verbatim}
\usepackage{graphicx}
\usepackage{mathtools} 
\usepackage{algorithm}        
\usepackage{algpseudocode}    
\usepackage{subcaption}  
\usepackage{booktabs} 
\usepackage{siunitx}  
\usepackage{multirow} 
\usepackage{amsthm}  

\theoremstyle{plain} 
\newtheorem{proposition}{Proposition}
\newtheorem{assumption}{Assumption}

\usepackage[style=ieee]{biblatex}
\begin{document}

\title{Land-then-transport: A Flow Matching-Based Generative Decoder for Wireless Image Transmission}

\author{Jingwen Fu,~\IEEEmembership{Student Member,~IEEE,}
Ming Xiao,~\IEEEmembership{Senior Member,~IEEE,}
Mikael Skoglund,~\IEEEmembership{Fellow,~IEEE,}
Dong In Kim,~\IEEEmembership{Life Fellow,~IEEE}

\thanks{Jingwen Fu, Ming Xiao, Mikael Skoglund are with the School of Electrical Engineering and Computer Science (EECS), KTH Royal Institute of Technology, 11428 Stockholm, Sweden. (Corresponding author: Ming Xiao.) Email: \{jingwenf, mingx, skoglund\}@kth.se.} 
\thanks{Dong In Kim is with the College of Information and Communication Engineering, Sungkyunkwan University, Suwon, South Korea. Email: dongin@skku.edu.}

}



\maketitle

\begin{abstract}
Due to stringent requirements on data rate and reliability, image transmission over wireless channels remains challenging for both classical layered designs and joint source–channel coding (JSCC), particularly under low-latency constraints. By leveraging powerful learned image priors, diffusion-based generative decoders can achieve strong perceptual quality under limited channel budgets. However, they normally have high decoding latency due to iterative stochastic denoising. To overcome this limitation and enable low-latency decoding, we propose a flow-matching (FM)-based generative decoder under a new \emph{land-then-transport} (LTT) paradigm, which tightly integrates the physical wireless channel into a continuous-time probability flow. We first construct a Gaussian smoothing path for AWGN channels whose noise schedule monotonically indexes the effective noise levels, and derive a closed-form analytical \emph{teacher} velocity field along this path. A deep neural-network based \emph{student} vector field is then trained via conditional flow matching (CFM), yielding a deterministic, channel-aware ordinary differential equation (ODE) decoder with complexity linear in the number of ODE steps; at inference time, it only requires an estimate of the effective noise variance to set the ODE initialization time. We further show that Rayleigh fading and MIMO channels can be converted, via linear MMSE equalization and singular-value-domain processing, into AWGN-equivalent channels with calibrated effective starting times (the time $t^\star$ on the Gaussian path whose noise level matches the effective channel noise). Thus the same probability path and trained velocity field of AWGN decoders can be reused for Rayleigh and MIMO channels without retraining. For a fixed number of complex channel uses per image, experiments on MNIST, Fashion-MNIST, and DIV2K over AWGN, Rayleigh, and MIMO channels demonstrate that the proposed decoder consistently outperforms JPEG2000+LDPC, DeepJSCC, and diffusion-based baselines, while achieving a favorable perceptual visual quality with as few as a small number of ODE steps. The results show that the proposed LTT framework provides a deterministic, physically interpretable, and computation-efficient solution for generative wireless image decoding for various channels.

\end{abstract}

\begin{IEEEkeywords}
Wireless communication, Image transmission, Diffusion Models, Flow Matching
\end{IEEEkeywords}

\section{Introduction}
Image transmission is one of the most important tasks in modern communication systems, with applications such as visual sensing, remote monitoring, and immersive communications. Classical image transmission schemes are typically layered, with source coding and channel coding optimized separately~\cite{gunduz2024jsccsurvey}. While layered methods are theoretically optimal in the asymptotic regime, they can be highly suboptimal in practical, finite blocklength settings, particularly under latency or complexity constraints. Moreover, separate source and channel coding methods are often highly sensitive to channel mismatch and difficult to adapt for varying channels~\cite{suto2025semanticimage}.

In contrast, deep joint source channel coding (DeepJSCC) has emerged as a powerful alternative that directly maps images to channel symbols and reconstructs them at the receiver using deep neural networks (DNNs), thereby avoiding separation between source and channel coding~\cite{bourtsoulatze2019deep}. Beyond purely discriminative encoders/decoders, recent efforts have incorporated generative models to further improve reconstruction quality and robustness. In particular, score-based diffusion models and related denoising diffusion frameworks have demonstrated strong performance in image synthesis and restoration~\cite{song2021score,pu2025art}, and have recently been adopted as learned priors or plug-in denoisers for wireless image transmission~\cite{pei2025ldmsemcom}.

\begin{figure}[t]
\centering
\includegraphics[width=0.9\linewidth]{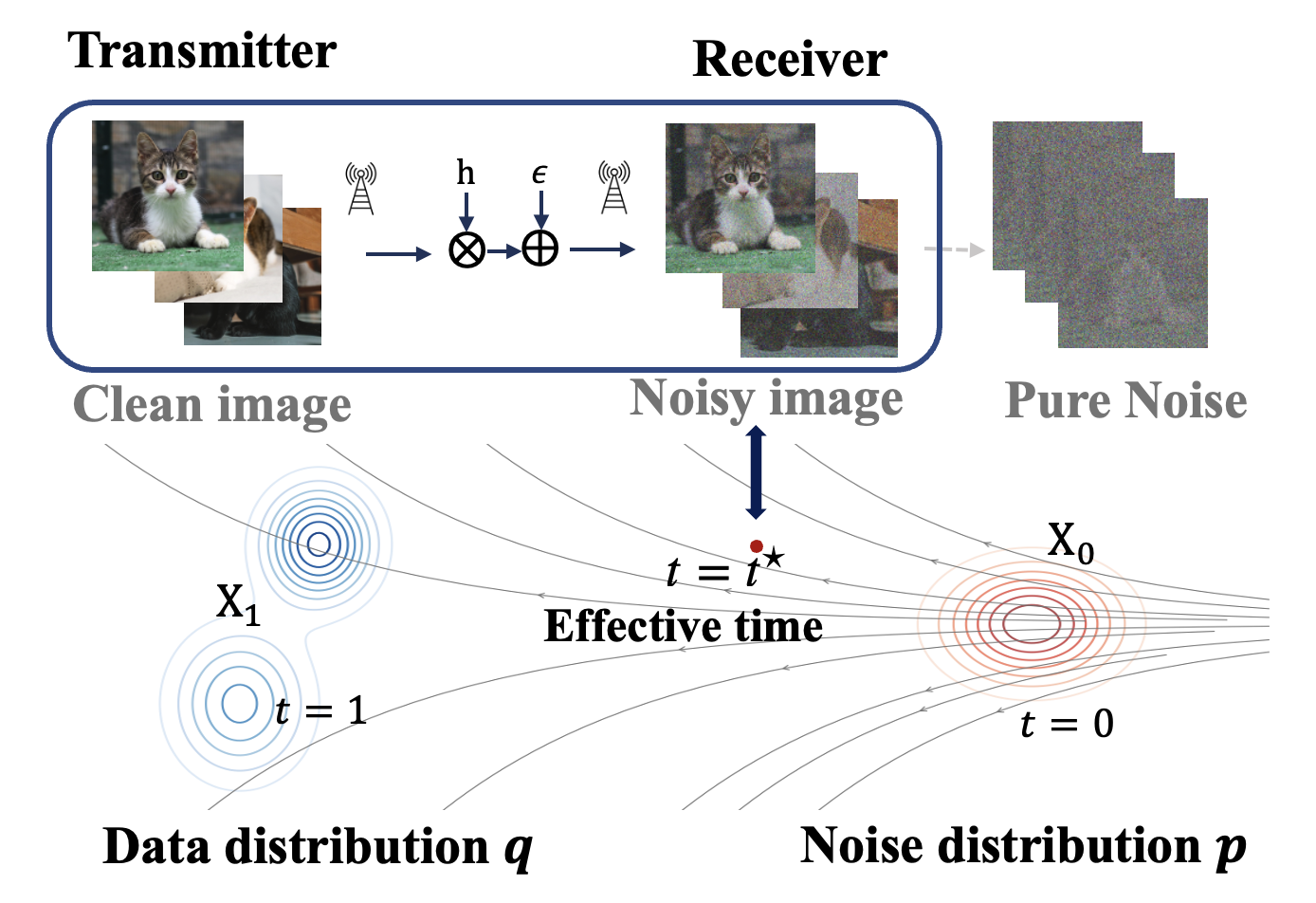}
\caption{Illustration of the proposed \emph{land-then-transport} scheme: channel outputs land on a Gaussian flow at an effective landing time $t^\star$, and are transported to clean images by a CFM-trained decoder shared across channels.}
\label{wire-flow}
\end{figure}

However, most existing generative decoders for wireless image transmission are diffusion-based and face two key limitations in communication settings~\cite{luong2025diffusion,fan2025generative}. First, diffusion-based receivers typically treat the wireless channel as an external source of corruption and apply a diffusion model to denoise the channel output, rather than embedding the channel effect into the generative dynamics~\cite{luong2025diffusion,cddm}. Consequently, the structural similarity between channel-noise corruption and diffusion noising/denoising is not fully exploited, and the resulting receivers are often multi-stage and over-parameterized. They use a large diffusion model as an additional module on top of the communication pipeline, which increases model complexity and training cost~\cite{cddm,guo2024diffsemcom}. Second, diffusion-based decoding relies on simulating a noisy forward process and a stochastic reverse process, which can make the training process fragile and is computationally demanding. Moreover, inference typically requires hundreds of reverse-diffusion steps~\cite{fan2025generative}, which incurs substantial sampling cost and decoding latency, and is particularly undesirable in low-latency or resource-constrained wireless systems.

Motivated by these observations, we adopt the recently proposed Flow Matching (FM) framework as the generative model for wireless image decoding~\cite{lipman2022flow}. FM learns a continuous-time, time-dependent velocity field that transports samples from a simple prior to the target data distribution along a prescribed probability path. Building on FM, Conditional Flow Matching (CFM) introduces analytically tractable conditional paths and leads to efficient regression-based training objectives~\cite{tong2024improving}. Leveraging these principles, we design a channel-aware decoder that tightly integrates the channel noise into the generative flow and replaces stochastic diffusion sampling with a deterministic ordinary differential equation (ODE)-based reconstruction procedure.
We propose a new \emph{land-then-transport} (LTT) paradigm that embeds wireless image transmission into a continuous-time generative flow. Specifically, we construct an FM probability path aligned with the physical channel, so that the received signal is interpreted as a noisy sample at an \emph{effective landing time $t^\star$} along the path. Decoding then reduces to transporting the landing point at $t^\star$ to the clean image distribution by solving a deterministic ODE, instead of running a long stochastic reverse diffusion process. The main contributions of this paper are summarized as follows.


\begin{itemize}
  \item To the best of our knowledge, we are the first to apply FM to end-to-end communication systems. We propose an LTT decoding paradigm that explicitly embeds the physical channel into a continuous-time probability flow. In our scheme, the channel output is interpreted as a landing point on the path at an effective landing time determined by the (effective) channel noise level, and decoding is carried out by solving a deterministic ODE from the landing point to the clean image distribution.
  
  \item For AWGN channels, we construct a Gaussian smoothing path whose \emph{noise schedule} (time-dependent mapping that specifies the noise level along the flow path) is aligned with the wireless channel, and derive a closed-form analytical \emph{teacher} velocity field along this path. We then instantiate a DNN-based \emph{student} vector field and train it via CFM to approximate the teacher field, which yields a single deterministic, channel-aware ODE decoder. The complexity of the decoder is determined by the number of ODE steps, and only the AWGN noise level at decoding is needed. In addition, we provide a scalar Gaussian channel analysis that characterizes the behavior of the decoder and the complexity–distortion trade-off for the ODE solver. 
  
\item Building on the results of AWGN channels, we show that Rayleigh fading and multi-input multi-output (MIMO) channels can be converted, via linear minimum mean square error (MMSE) equalization and singular value domain (SVD), into observations equivalent to AWGN channels with calibrated effective noise levels. Thus, we have a unified LTT decoder in which the same Gaussian path and AWGN-trained student velocity field can be reused for different channel models without retraining.

\item We conduct extensive experiments on MNIST, Fashion-MNIST, and DIV2K datasets over AWGN, Rayleigh, and MIMO channels. The proposed decoder consistently improves various performance metrics over  JPEG2000+LDPC, DeepJSCC, and diffusion-based generative baselines. We also achieve a deterministic, interpretable, and computation-efficient decoding process with a favorable visual perceptual quality. For example, on DIV2K dataset over AWGN channels at $\mathrm{SNR}=20$~dB, our method improves peak signal-to-noise ratio (PSNR, a distortion metric where higher indicates smaller reconstruction error) by $26.6\%$ and $28.3\%$ over a diffusion-based generative baseline and DeepJSCC, respectively, and increases multi-scale structural similarity index (MS-SSIM, a perceptual similarity metric where higher indicates better visual fidelity) by $53.2\%$ and $59.6\%$, and require only $10$ ODE steps at the decoder.
\end{itemize}

The remainder of this paper is organized as follows. Section~\ref{section:literature} reviews recent advances in wireless image transmission and diffusion-based generative decoding, and summarizes FM and CFM framework used in this work.
Section~\ref{sec:model} introduces the system model and formulates the proposed LTT decoding paradigm. Section~\ref{sec:proposed} details the AWGN Gaussian smoothing path, the CFM training and ODE-based decoding procedures, and provides theoretical analysis. Section~\ref{sec:rayleigh-mimo} extends the proposed framework to Rayleigh fading and MIMO channels. Section~\ref{sec:results} presents numerical results and ablation studies. Finally, Section~\ref{sec:conclusion} concludes the paper.

Notations: Random variables are denoted by uppercase letters (e.g., $X$) and their realizations by lowercase letters (e.g., $x$). Boldface lowercase (e.g., $\mathbf x$) and uppercase (e.g., $\mathbf H$) denote vectors and matrices, respectively, and $\mathbf I$ denotes the identity matrix. For real and complex scalars, $|\cdot|$ denotes the modulus, while for vectors $\|\cdot\|$ denotes Euclidean norm. $(\cdot)^{\mathsf T}$ and $(\cdot)^{\mathsf H}$ denote transpose and Hermitian transpose operators, respectively, and $(\cdot)^{\ast}$ denotes complex conjugation. Real and circularly symmetric complex Gaussian distributions with mean $\mu$ and covariance $\Sigma$ are denoted as $\mathcal N(\mu,\Sigma)$ and $\mathcal{CN}(\mu,\Sigma)$, respectively, and $\mathbb E[\cdot]$ denotes expectation. $\mathbb R^d$ and $\mathbb C^d$ are the $d$-dimensional real and complex vector spaces.

\section{Related Work}
\label{section:literature}
In this section, we provide a review of recent work on image transmission in wireless systems. Then, we will briefly review diffusion, FM, and CFM models.

\subsection{Image Transmission in Wireless Systems}

Classical image transmission typically follows the source--channel separation paradigm, where images are first compressed (e.g., JPEG, BPG) and then protected by channel coding~\cite{ETSI_EN_302_755}. While asymptotically optimal, such layered schemes suffer from the cliff effect and are fragile under channel mismatch or stringent latency and bandwidth constraints~\cite{986011}. DeepJSCC addresses the limitations by learning an end-to-end mapping from pixels to complex channel symbols via convolutional autoencoders, thereby mitigating the cliff effect and outperforming layer-based schemes, especially in the low-SNR and low-bandwidth regimes~\cite{bourtsoulatze2019deepjscc}. Several follow-up works have extended DeepJSCC to MIMO channels~\cite{wu2024deepjsccmimo} and resource-adaptive architectures under computational and bandwidth budgets~\cite{fu2025computation}.

More recently, generative models have been proposed for image transmission to further enhance perceptual quality. Representative approaches include diffusion-based denoisers after channel equalization~\cite{cddm}, diffusion-driven semantic communication with compression~\cite{guo2024diffsemcom}, latent diffusion with end-to-end consistency distillation for few-step denoising~\cite{pei2025ldmsemcom}, semantic-guided diffusion for DeepJSCC~\cite{zhang2025sgdjscc}, and diffusion-enabled semantic schemes that transmit highly compressed cues such as edge maps~\cite{grassucci2023generative}, etc. While these approaches demonstrate the potential of generative priors for image communications, they typically rely on large diffusion backbones with many sampling steps and often treat the physical channel as an external source of noise, rather than explicitly embedding the channel effect into the generative process. This motivates the development of lightweight, channel-aware generative decoders for wireless communication.

\begin{figure*}[t]
    \centering 
    \subfloat[Flow]{\includegraphics[width=0.24\textwidth]{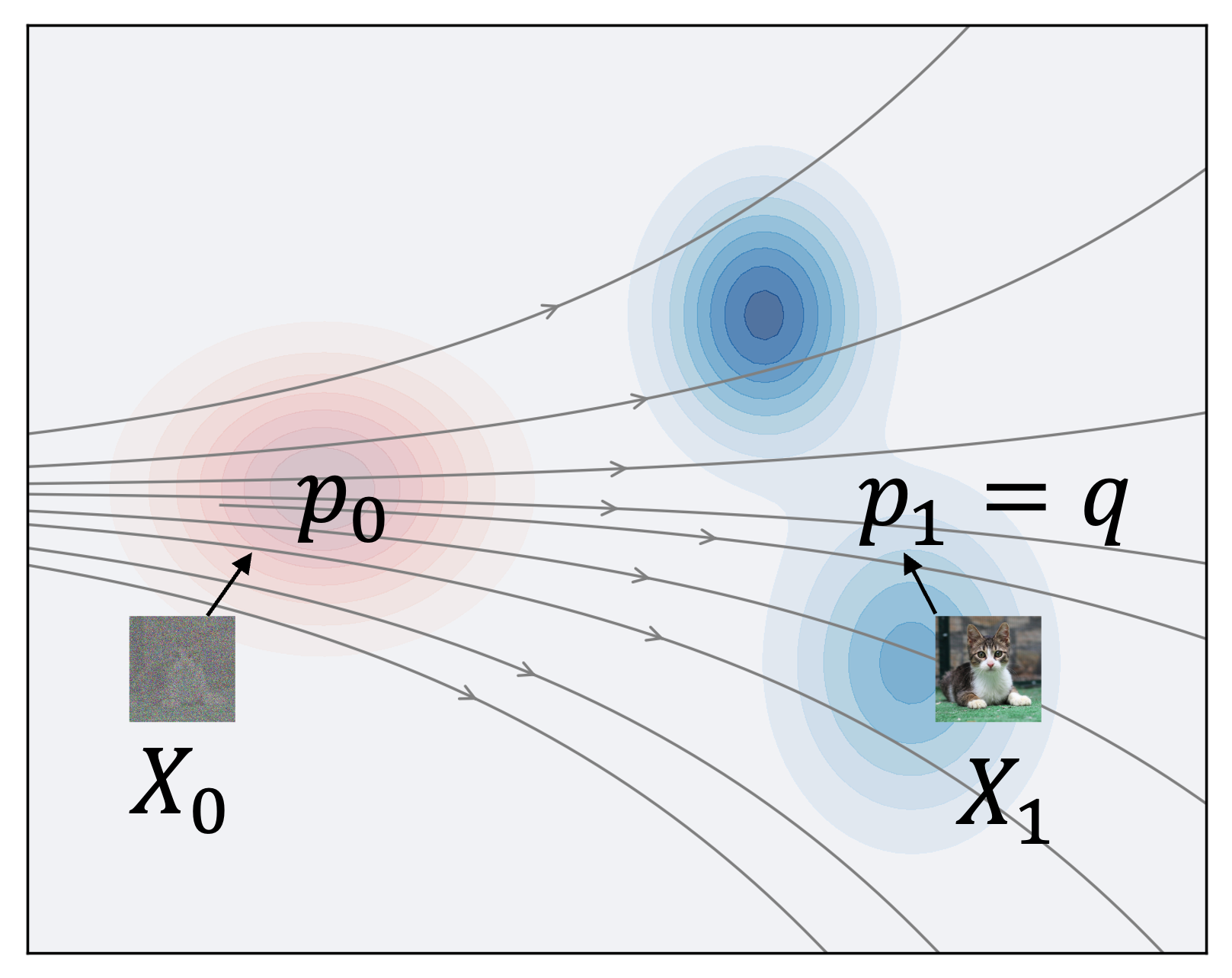}
        \label{fig:sub1}}
    \hfill 
    \subfloat[Path]{\includegraphics[width=0.24\textwidth]{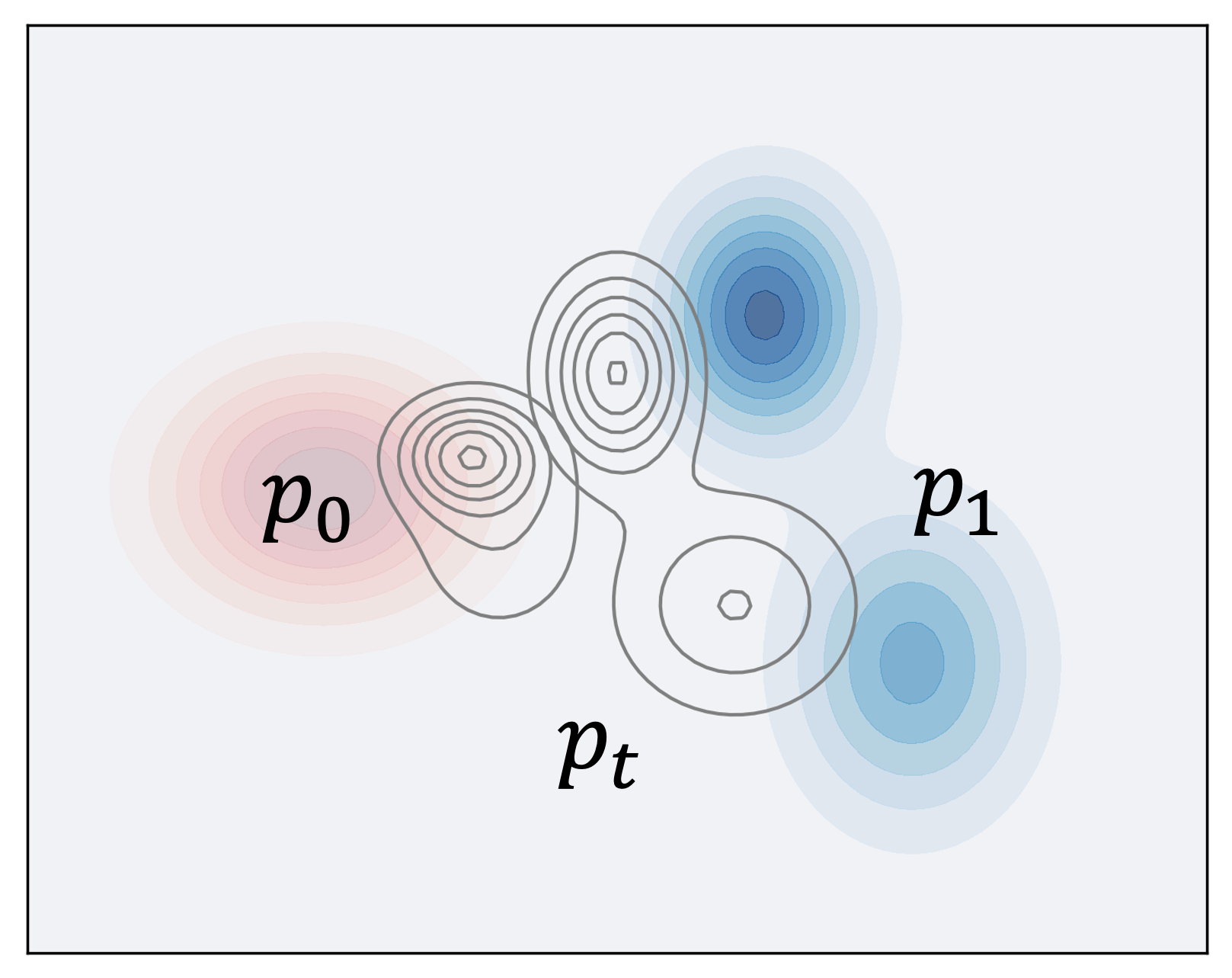}
        \label{fig:sub2}}
    \hfill 
    \subfloat[Training]{\includegraphics[width=0.24\textwidth]{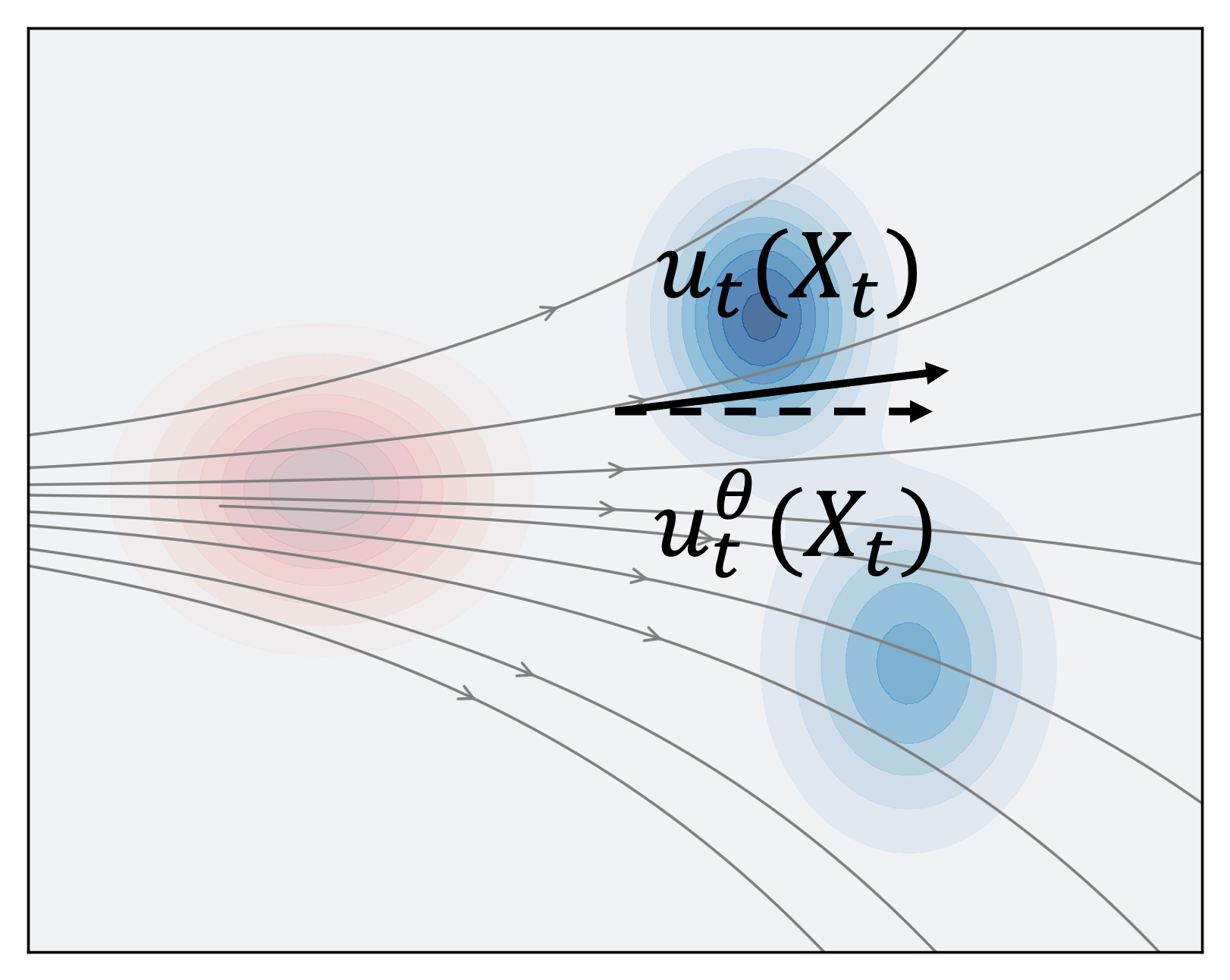}
        \label{fig:sub3}}
    \hfill
    \subfloat[Conditional path]{\includegraphics[width=0.24\textwidth]{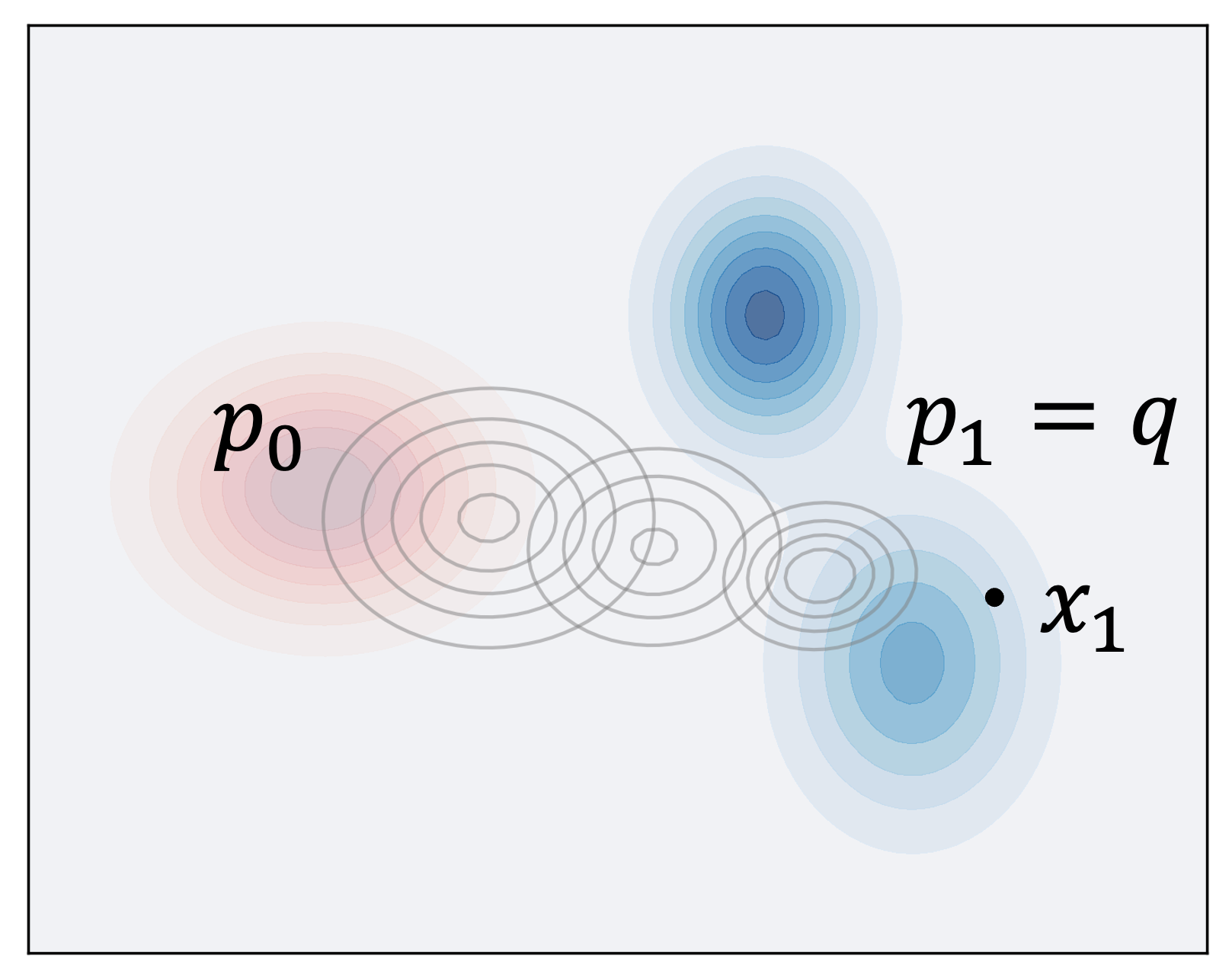}
        \label{fig:sub4}}
    \caption{Illustration of FM and CFM.
(a) A velocity field transports samples from a simple prior $p_0$ to a target distribution $q$ along continuous trajectories.
(b) The induced probability path $(p_t)_{t \in [0,1]}$ smoothly interpolates between $p_0$ and $p_1 = q$.
(c) FM trains a neural velocity field $u_t^\theta(X_t)$ to match the true velocity $u_t(X_t)$ along this path.
(d) CFM replaces the intractable marginal path with a tractable conditional linear path from $p_0$ to $p_1=q$ by conditioning on $X_1 = x_1$.}
\label{fig:fm_concept}
\end{figure*}

\subsection{Diffusion Models and FM Models}

Diffusion and score-based generative models have been recently proposed for image generation by progressively corrupting data with Gaussian noise and learning to reverse this process. Denoising diffusion probabilistic models (DDPMs) discretize the forward noising process into a Markov chain and train a network to predict the clean sample at each step~\cite{ho2020denoising}. Score-based models learn the score function of the noisy intermediate distribution, i.e., the gradient of the log-density with respect to the data,
and realize generation as the solution of a reverse-time stochastic differential equation that transforms a simple prior to the data distribution~\cite{song2021score}. Acceleration techniques such as deterministic samplers, probability flow ODEs, and consistency models reduce the number of reverse steps~\cite{song2023consistency}. However, these methods still approximate the reverse dynamics of an underlying stochastic process and remain computationally demanding at inference.

Flow-based generative models, including continuous normalizing flows, instead parameterize generation directly as the solution of a deterministic ODE driven by a neural vector field. FM learns this vector field by regressing it to an analytically specified velocity along a prescribed probability path, thereby avoiding explicit simulation of a forward diffusion process~\cite{lipman2022flow}. Building on FM, CFM introduces conditional velocity fields depending on both the current state and data, expressing the marginal field as expectation over tractable conditional flows and yielding a unified framework that links flows and diffusion models~\cite{tong2024improving}. However, these efforts have been explored mainly in generic generative modeling settings, and it remains unclear how to exploit FM as a channel-aware decoder whose probability path is aligned with the noise statistics of practical AWGN, Rayleigh, and MIMO channels. Motivated by this gap, we develop an FM-based generative decoder for wireless image transmission using a channel-aligned Gaussian smoothing path.


\section{System Model and Problem Formulation}
\label{sec:model}

In this section, we will briefly review FM and CFM, and then propose LTT method, in which wireless image transmission is treated as one step on the flow path.

\subsection{Preliminaries on FM and CFM}
\label{sec:preliminary}

Given a training dataset of samples from a targeted distribution $q$ over $\mathbb{R}^d$, the goal of a generative model is to approximate the distribution, from which new samples will be generated~\cite{lipman2022flow}. FM achieves the objective by introducing a continuous-time probability path $(p_t)_{t \in [0,1]}$ that smoothly interpolates between a simple prior distribution and the data distribution. Specifically, the path starts from a prior $p_{t=0} = p_0$ (e.g., $p_0 = \mathcal{N}(0, I_d)$) and ends at the target distribution $p_{t=1} = p_1 = q$. As illustrated in Fig.~\ref{fig:fm_concept}(a), the evolution of $t$ from $0$ to $1$ can be viewed as a family of trajectories that continuously transport probability mass from $p_0$ at time $t=0$ to $q$ at time $t=1$.

Formally, FM specifies a time-dependent velocity field $u : [0, 1] \times \mathbb{R}^d \to \mathbb{R}^d$ that governs the evolution of particles along the path. The velocity field induces a flow of diffeomorphisms $\{\psi_t\}_{t \in [0,1]}$ through the ODE
\begin{equation}
    \frac{\mathrm{d}}{\mathrm{d}t} \psi_t(x) = u_t(\psi_t(x)), \qquad \psi_0(x) = x.
\end{equation}
If $X_0 \sim p_0$ and we define $X_t = \psi_t(X_0)$, then $X_t$ is distributed according to $p_t$, such that $p_t$ traces the desired transport from $p_0$ at $t=0$ to $q$ at $t=1$.
 Fig.~\ref{fig:fm_concept}(b) depicts how the path $p_t$ interpolates between the prior and the data distribution.

In practice, the unknown velocity field $u_t$ is represented by a neural network $u_t^\theta$ with parameters $\theta$. The goal of FM is to estimate $\theta$ by minimizing the expected squared error between $u_t^\theta$ and the true velocity $u_t$
over $t$ and samples $X_t \sim p_t$, i.e., 
\begin{equation}
    L_{\text{FM}}(\theta)
    = \mathbb{E}_{t,\, X_t \sim p_t}
      \big[ \| u_t^\theta(X_t) - u_t(X_t) \|^2 \big],
    \label{eq:fm_loss}
\end{equation}
where $t \sim \mathcal{U}[0,1]$. As shown in Fig.~\ref{fig:fm_concept}(c), FM training updates $u_t^\theta$ so that the predicted velocity (solid arrow) aligns with the ground-truth velocity (dashed arrow) that transports $X_t$ along the flow. However, the objective is generally intractable, since both the marginal distributions $p_t$ and the true velocity field $u_t$ are unknown.

CFM~\cite{tong2024improving} circumvents the difficulty by introducing a specific, tractable probability path known as the conditional optimal-transport path. To sample $X_t \sim p_t$ on the path, one first draws $X_0 \sim p_0$ and $X_1 \sim q$, and then linearly interpolates between them:
\begin{equation}
    X_t = (1-t) X_0 + t X_1.
    \label{eq:linear_path}
\end{equation}
For each fixed endpoint $x_1$, the conditional trajectory $t \mapsto X_t \mid X_1 = x_1$ is a straight line connecting $x_0$ and $x_1$, as illustrated in Fig.~\ref{fig:fm_concept}(d). The corresponding conditional velocity field that generates this path follows a closed form:
\begin{equation}
    u_t(x \mid x_1) = \frac{x_1 - x}{1-t}.
\end{equation}

The closed-form expression enables a tractable training objective, the CFM loss, which regresses the neural velocity field towards the known conditional velocity:
\begin{equation}
    L_{\text{CFM}}(\theta)
    = \mathbb{E}_{t,\, X_0 \sim p_0,\, X_1 \sim q}
      \big[ \| u_t^\theta(X_t) - u_t(X_t \mid X_1) \|^2 \big],
    \label{eq:cfm_loss}
\end{equation}
where $X_t$ is given by \eqref{eq:linear_path}. Remarkably, although $L_{\text{CFM}}$ is defined conditionally on $X_1$, it yields the same gradient as the original FM objective as \eqref{eq:fm_loss}~\cite{lipman2022flow}
\begin{equation}
    \nabla_\theta L_{\text{FM}}(\theta)
    = \nabla_\theta L_{\text{CFM}}(\theta).
    \label{eq:grad_equiv}
\end{equation}
Therefore, we can efficiently train $u_t^\theta$ by minimizing $L_{\text{CFM}}$ while still optimizing the original FM objective.

\subsection{LTT: Wireless Transmission as One Step of the Flow Path}
\label{sec:land-then-transport}

We now embed the wireless channel into the probability path $\{p_t\}_{t\in[0,1]}$ defined above. As illustrated in Fig.~\ref{wire-flow}, let $p(x)=\mathcal{N}(0,I_d)$ denote a simple source prior and let $q(x)$ be the clean data distribution on $\mathbb{R}^d$, with $X_0\!\sim\!p$ and $X_1\!\sim\!q$.
We first consider an AWGN channel where the transmitter sends a clean image $X_1\!\sim\!q$ and the receiver observes
\begin{equation}
    Y = X_{1} + \sigma_{\mathrm{ch}}\varepsilon,
    \qquad \varepsilon \sim \mathcal N(0,I_d).
    \label{eq:awgn-channel}
\end{equation}
Our goal is to choose a probability path and a time-dependent noise schedule such that the strength of the corruption along the path is monotonically indexed by the time variable~$t$. Intuitively, one can view $t$ as a continuous SNR index: small $t$ corresponds to low SNR (strong noise), while $t=1$ corresponds to the clean end point, i.e., samples from the data distribution without added noise. 
More formally, we introduce a strictly decreasing \emph{noise schedule}
\begin{equation}
  \sigma(t):[0,1]\to[0,\sigma_{\max}],\qquad \sigma(0)=\sigma_{\max},\ \sigma(1)=0,
  \label{eq:sigma-schedule}
\end{equation}
where $\sigma_{\max}$ is chosen to upper-bound the channel noise levels of interest (i.e., $\sigma_{\max}\ge \sigma_{\mathrm{ch}}$).
Since $\sigma(t)$ is strictly monotone (and we take it continuous), it is a bijection between $[0,1]$ and $[0,\sigma_{\max}]$, and hence the inverse mapping $\sigma^{-1}:[0,\sigma_{\max}]\to[0,1]$ is well-defined. Therefore, for any admissible channel noise level $\sigma_{\mathrm{ch}}\in(0,\sigma_{\max}]$, we can map it to a unique \emph{effective landing time}
\begin{equation}
    t^{\star} = \sigma^{-1}(\sigma_{\mathrm{ch}}).
    \label{eq:tstar-awgn}
\end{equation}

Thus, the channel output $Y$ has the same conditional distribution as the path state at $t^\star$. That is, $Y\,|\,X_1=x_1$ is distributed as $X_{t^\star}\,|\,X_1=x_1$. Thus, we can interpret $Y$ as a realization of $X_{t^\star}$ lying on the flow path at time $t^\star$.

At the receiver, we first compute $t^{\star}$ via~\eqref{eq:tstar-awgn} and identify the path state with the channel output, setting $X_{t^{\star}}=Y$. We then integrate a learned probability-flow ODE
\begin{equation}
    \frac{\mathrm{d}}{\mathrm{d}t} X_t = v_{t}^{\theta}(X_t), 
    \qquad t \in [t^\star, 1], \quad X_{t^\star} = Y,
    \label{eq:ltt-ode}
\end{equation}
forward in time from $t^{\star}$ to $t=1$ to obtain the estimate $\widehat{X}_{1}$. We refer to the process as the LTT decoding strategy: the physical first channel \emph{lands} the signal at time $t^\star$ on the path, and the learned flow deterministically \emph{transports} it to the clean endpoint.
Under the view, the overall decoding process can be summarized as follows. In the \emph{offline training phase}, we fix a noise schedule $\sigma(t)$ and train a parametric vector field $v_t^{\theta}$ along the path $\{p_t\}$ using CFM. In the \emph{online decoding phase}, for each SNR the receiver maps the channel noise level $\sigma_{\mathrm{ch}}$ to an effective landing time $t^\star$, and identifies $X_{t^\star}$ with the received $Y$, and integrates~\eqref{eq:ltt-ode} from $t^\star$ to $1$ to reconstruct the clean image. The detailed construction of the Gaussian path, the associated analytic velocity field, and the CFM training objective for the DNN-based field $v_t^\theta$ will be given in Sec.~\ref{sec:proposed}.

\section{Proposed Method in AWGN Channels}
\label{sec:proposed}

In this section, we instantiate the proposed LTT framework for  real-valued AWGN channels in detail. We first construct an AWGN-compatible flow path and its DNN-based student velocity field together with the corresponding CFM training algorithm. We then describe the decoding procedure at the receiver and provide theoretical analysis. 

\subsection{AWGN Channel Flow Path} 
\label{subsec:awgn-path}

\begin{table}[t]
    \centering
    \caption{Architecture of the proposed U-Net student velocity field network.}
    \label{tab:unet-arch}
    \setlength{\tabcolsep}{3pt}
    \begin{tabular}{c|c|c|p{3.5cm}}
        \hline
        Stage & Level & Channels & Main operations \\
        \hline
        Input  & --   & $C_{\text{in}}\!\times\!H\!\times\!W$ & Input, time embedding \\
        \hline
        \multirow{4}{*}{Encoder}
          & 0 & 64  & Conv $3\times3$, $2\times$ ResBlock, Downsample \\
          & 1 & 128 & $2\times$ ResBlock, Downsample \\
          & 2 & 256 & $2\times$ ResBlock, Attention, Downsample \\
          & 3 & 512 & $2\times$ ResBlock, Attention \\
        \hline
        Middle & -- & 512 & ResBlock, Attention, ResBlock \\
        \hline
        \multirow{4}{*}{Decoder}
          & 3 & 512 & Concat skip-3, $3\times$ ResBlock, Attention, Upsample \\
          & 2 & 256 & Concat skip-2, $3\times$ ResBlock, Attention, Upsample \\
          & 1 & 128 & Concat skip-1, $3\times$ ResBlock, Upsample \\
          & 0 & 64  & Concat skip-0, $3\times$ ResBlock \\
        \hline
        Output & -- & $C_{\text{out}}\!\times\!H\!\times\!W$ & GroupNorm, SiLU, Conv $3\times3$ \\
        \hline
    \end{tabular}
\end{table}

Building on the LTT formulation in Sec.~\ref{sec:land-then-transport}, we can create the Gaussian smoothing path and its generating velocity field explicit for real-valued AWGN channels. Recall that $X_{1}\sim q(x)$ denotes the clean data and that the received signal is given by an AWGN channel model in~\eqref{eq:awgn-channel}. We reuse the strictly decreasing noise schedule $\sigma(t)$ introduced in~\eqref{eq:sigma-schedule} and specify an AWGN-compatible conditional flow path with the mean anchored at $x_{1}$ while the modified variance with $t$:
\begin{equation}
X_t\,\big|\,X_1=x_1 \sim \mathcal N\!\big(x_1,\;\sigma(t)^2 I_d\big),\quad t\in[0,1].
\label{eq:awgn-path}
\end{equation}
The induced marginal path is the Gaussian smoothing of $q$ as
\begin{equation}
\label{eq:pt-marginal}
\begin{aligned}
p_t(x)
&= \int \mathcal N\big(x;\,x_1,\sigma(t)^2 I_d\big)\,q(x_1)\,\mathrm{d}x_1 \\
&= \big(q*\mathcal N(0,\sigma(t)^2 I_d)\big)(x).
\end{aligned}
\end{equation}
By the definition of the landing time $t^\star$ in~\eqref{eq:tstar-awgn}, the channel output $Y$ lies exactly on path $p_t(x)$, in the sense that $Y\,|\,X_1 \overset{d}{=} X_{t^\star}\,|\,X_1$.
By~\cite{lipman2022flow}, the conditional vector field 
$u_t(\cdot \mid x_1) : \mathbb{R}^d \to \mathbb{R}^d$  generating $p_t(\cdot \mid x_1)$ has the form
\begin{equation}
    u_t(x \mid x_1)
    = \frac{\dot{\sigma}(t)}{\sigma(t)} \bigl(x - \mu_t(x_1)\bigr)
      + \dot{\mu}_t(x_1),
\end{equation}
where $\dot{\sigma}(t) = \tfrac{\mathrm{d}}{\mathrm{d}t}\sigma(t)$ and 
$\dot{\mu}_t(x_1) = \tfrac{\mathrm{d}}{\mathrm{d}t}\mu_t(x_1)$.
For our AWGN flow path, we set $\mu_t(x_1) = x_1$, thus $\dot{\mu}_t(x_1)=0$ and
\begin{equation}
    u_t(x \mid x_1)
    = \frac{\dot{\sigma}(t)}{\sigma(t)}\,(x - x_1),
\end{equation}
which can be interpreted as a homogeneous contraction toward $x_1$. To verify a velocity field $u_t$ generating a probability path $p_t$, one can check pair $(u_t, p_t)$ satisfying Continuity Equation:
\begin{equation}
    \frac{\mathrm{d}}{\mathrm{d}t} p_t(x) + \operatorname{div}\big(p_t u_t\big)(x) = 0,
\end{equation}
where $\operatorname{div}(v)(x) = \sum_{i=1}^{d} \partial_{x^i} v^i(x)$ for $v(x) = \big(v^1(x), \ldots, v^d(x)\big)$. The detailed proof is given in Appendix~\ref {app:continuity}. Therefore, having shown that our proposed velocity field $u_t$ generates the desired probability path $p_t$, we can train a neural vector field $v_{\theta}(x,t)$ by CFM using the regression loss
\begin{equation}
\label{eq:cfm-loss-2col}
\begin{aligned}
\mathcal{L}_{\mathrm{CFM}}(\theta)
&= \mathbb{E}_{t\sim U[0,1]}\,\mathbb{E}_{x_{1}\sim q}\,\mathbb{E}_{x\sim p_{t}(\cdot\mid x_{1})} \\
&\quad \Bigl\| v_{\theta}(x,t) - u_{t}(x\mid x_{1}) \Bigr\|^{2}.
\end{aligned}
\end{equation}
As stated in Sec~\ref{sec:preliminary}, we have $\nabla_{\theta}\mathcal{L}_{\mathrm{CFM}}(\theta)=\nabla_{\theta}\mathcal{L}_{\mathrm{FM}}(\theta)$.


\subsection{Student Velocity Field Network}

Given the analytical conditional field $u_t(\cdot \mid x_1)$ in~\eqref{eq:awgn-path}–\eqref{eq:cfm-loss-2col}, we learn a parametric \emph{student} velocity field, i.e., a neural approximation
\begin{equation}
    v_{\theta}(x,t) \approx u_t(x \mid x_1),
\end{equation}
where $v_{\theta}:\mathbb{R}^{C\times H\times W}\times[0,1]\to\mathbb{R}^{C\times H\times W}$ is the velocity-field function implemented by a neural network, and $\theta$ denotes its trainable weights. 
The student field is trained to approximate the analytical \emph{teacher} field $u_t(\cdot\mid x_1)$, enabling efficient inference.
For image data, each state is an image tensor $x_t \in \mathbb{R}^{C \times H \times W}$,
and both the teacher field $u_t(x_t \mid x_1)$ and the student field $v_{\theta}(x_t,t)$ output a velocity tensor in the same space $\mathbb{R}^{C \times H \times W}$.
We implement $v_{\theta}$ as a U-Net convolutional network~\cite{ronneberger2015u}. Given $(x,t)$, it produces
\begin{equation}
    \hat{u} = v_{\theta}(x,t) \in \mathbb{R}^{C \times H \times W}.
\end{equation}
The structure of $v_{\theta}$ is summarized in Table~\ref{tab:unet-arch}. The backbone is a standard encoder–decoder U-Net with residual blocks, down/upsampling, and self-attention at deeper layers, providing multi-scale spatial features~\cite{ronneberger2015u}.


\subsection{Training the Student Velocity Field}

The learning objective is CFM  loss in \eqref{eq:cfm-loss-2col}, which drives the student field $v_{\theta}$ to match the teacher field $u_t(\cdot \mid x_1)$ along the AWGN path. At each training step, we sample a clean target $x_1$, draw a random time $t$ and Gaussian noise, construct an intermediate state $x_t$ on the path, evaluate the closed-form teacher velocity $u_t(x_t \mid x_1)$, and regress the student prediction $v_{\theta}(x_t,t)$ onto the target  $u_t(x_t \mid x_1)$. The training procedure is summarized in Algorithm~\ref{alg:fm-train}.

\begin{algorithm}[t]
\caption{Training procedure of the proposed LTT decoder}
\label{alg:fm-train}
\textbf{Input} Training set $\mathcal{D}_{\text{train}}$, epochs $T$, batch size $B$, noise schedule $\sigma(\cdot)$ \\
\textbf{Output} Trained parameters $\theta$ of the student field $v_{\theta}$ \\
\textbf{Initialization} Initialize $\theta$
\begin{algorithmic}[1]
\For{$\text{epoch} = 1$ to $T$}
    \State Sample a mini-batch $\{x_1^{(i)}\}_{i=1}^{B} \subset \mathcal{D}_{\text{train}}$
    \State Sample $t^{(i)} \sim \mathcal{U}[0,1]$ and $\varepsilon^{(i)} \sim \mathcal{N}(0,I)$
    \State $\sigma^{(i)} \gets \sigma\bigl(t^{(i)}\bigr)$, \quad $\dot{\sigma}^{(i)} \gets \dot{\sigma}\bigl(t^{(i)}\bigr)$ 
        \Comment{Evaluate schedule and its derivative}
    \State $x_t^{(i)} \gets x_1^{(i)} + \sigma^{(i)} \varepsilon^{(i)}$ 
        \Comment{Sample along AWGN path}
    \State $u^{(i)} \gets \dfrac{\dot{\sigma}^{(i)}}{\sigma^{(i)}}\bigl(x_t^{(i)} - x_1^{(i)}\bigr)$
        \Comment{Teacher velocity }
    \State $\hat{u}^{(i)} \gets v_{\theta}\bigl(x_t^{(i)}, t^{(i)}\bigr)$
        \Comment{Student prediction }
    \State $\widehat{\mathcal{L}}_{\mathrm{CFM}} \gets \dfrac{1}{B}\sum_{i=1}^{B} \bigl\|\hat{u}^{(i)} - u^{(i)}\bigr\|^{2}$
    \State Update $\theta$ by gradient descent on $\widehat{\mathcal{L}}_{\mathrm{CFM}}$
\EndFor
\end{algorithmic}
\end{algorithm}

\subsection{Decoding at the Receiver}

\begin{figure}[t]
    \centering
    \subfloat[FM \label{result-dynamic-awgn}]{
        \includegraphics[width=0.85\linewidth]{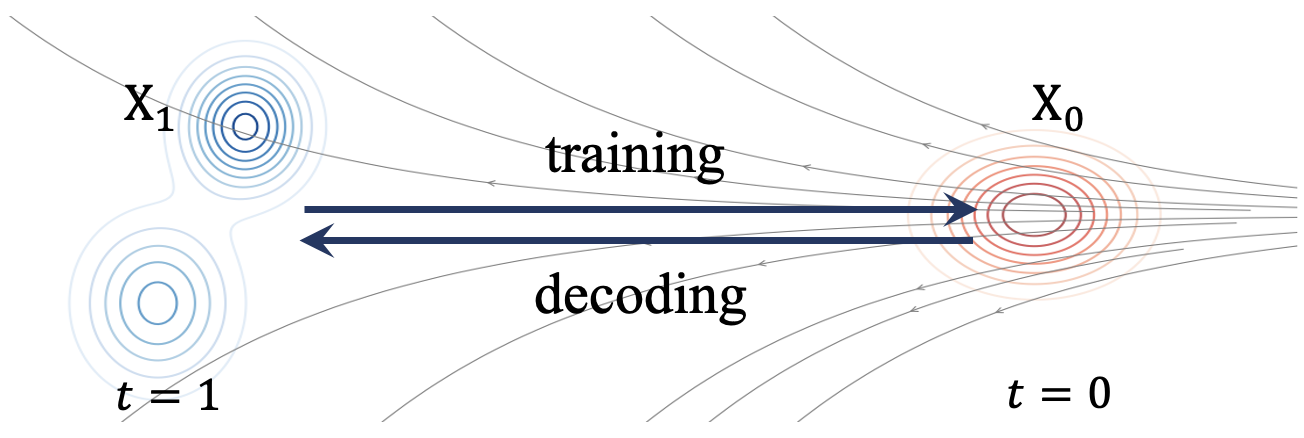}
    }\\
    \subfloat[LTT \label{result-dynamic-rayleigh}]{
        \includegraphics[width=0.85\linewidth]{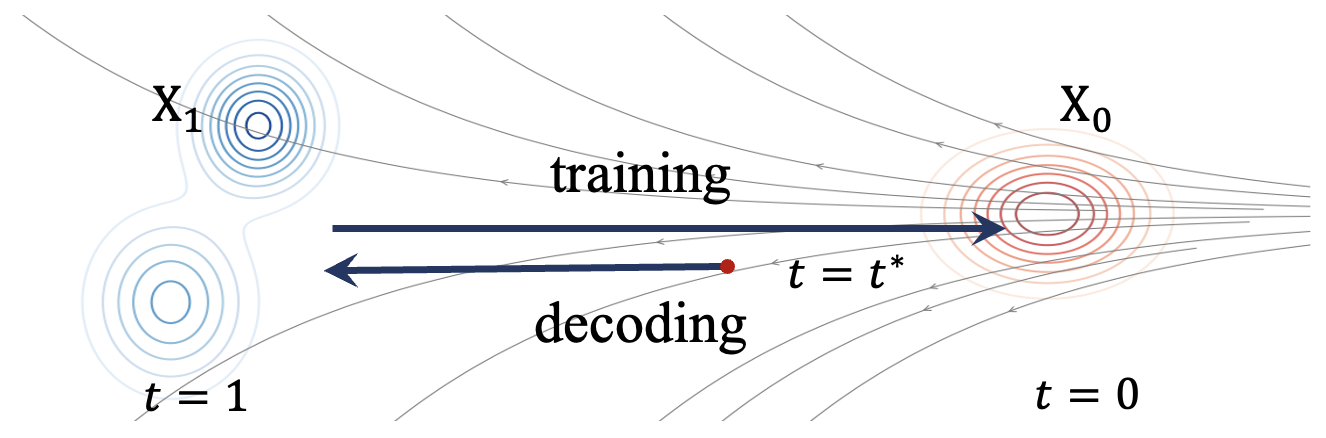}
    }
    \caption{Training and decoding path in FM and the proposed LTT method.}
    \label{fig:flow-and-ours}
\end{figure}

Given a trained student field $v_{\theta}$ and a fixed noise schedule $\sigma(t)$, decoding under an AWGN channel follows directly from the continuous-time formulation in~\eqref{eq:ltt-ode}. For a wireless channel with AWGN noise variance $\sigma_{\mathrm{ch}}^{2}$ and received observation $Y$, the receiver first computes the landing time
\(
t^{\star} = \sigma^{-1}(\sigma_{\mathrm{ch}})
\)
according to~\eqref{eq:tstar-awgn} and sets the initial state on the flow as $X_{t^{\star}} = Y$. The remaining task is to solve the probability-flow ODE~\eqref{eq:ltt-ode} forward from $t= t^{\star}$ to $t=1$ to obtain the reconstruction $\widehat X_1$. As shown in Fig.~\ref{fig:flow-and-ours}, different from standard FM sampling, which starts from $X_0 \sim p_0$ at $t=0$ and integrates over the entire interval $[0,1]$, the proposed LTT decoder only integrates over $[t^{\star},1]$, with the interval $[0,t^{\star}]$ effectively realized by the physical channel. By replacing the early part of the flow with the wireless channel, the decoder  preserves the FM generative structure while explicitly incorporates the wireless channel as a part of the probability path.
In practice, we discretize the interval $[t^{\star},1]$ into $N$ uniform steps with step size
\begin{equation}
    \Delta t = \frac{1 - t^{\star}}{N},
\end{equation}
and approximate the ODE solution using a standard numerical solver. With first-order Euler method, updates at the $k$-th step are
\begin{equation}
    x_{t_{k+1}} = x_{t_k} + \Delta t\,v_{\theta}(x_{t_k}, t_k), 
    \qquad t_k = t^{\star} + k \Delta t,
\end{equation}
for $k = 0,\dots,N-1$. As a simple higher-order alternative, we also consider the second-order midpoint method,
\begin{equation}
    x_{t_{k+1}} 
    = x_{t_k} 
      + \Delta t\,v_{\theta}\!\Big(x_{t_k} + \tfrac{\Delta t}{2}\,v_{\theta}(x_{t_k}, t_k),\; t_k + \tfrac{\Delta t}{2}\Big),
\end{equation}
which reduces integration errors while keeping the decoding process fully deterministic. The LTT decoding algorithm is summarized in Algorithm~\ref{alg:fm-decode}.


\subsection{Scalar Gaussian Benchmark}

\begin{algorithm}[t]
\caption{Decoding of the proposed LTT decoder}
\label{alg:fm-decode}
\textbf{Input} Received $y$, channel noise variance $\sigma_{\mathrm{ch}}^{2}$, schedule $\sigma(\cdot)$, trained $v_{\theta}$, steps $N$ \\
\textbf{Output} Reconstructed image $\widehat{x}_1$ \\
\textbf{Initialization} $t^{\star} \gets \sigma^{-1}(\sigma_{\mathrm{ch}})$, \quad $\Delta t \gets (1 - t^{\star})/N$
\begin{algorithmic}[1]
    \State $x^{(0)} \gets y$, \quad $t_0 \gets t^{\star}$
    \For{$k = 0$ to $N-1$}
        \State $v^{(k)} \gets v_{\theta}\bigl(x^{(k)}, t_k\bigr)$
        \State $t_{k+1} \gets t_k + \Delta t$
        \State $x^{(k+1)} \gets x^{(k)} + \Delta t\, v^{(k)}$ \Comment{Euler / Midpoint}
    \EndFor
    \State $\widehat{x}_1 \gets x^{(N)}$
\end{algorithmic}
\end{algorithm}

To gain further insights into the proposed LTT decoder, we consider a simplified scalar Gaussian wireless channel setting
\begin{equation}
    X_1 \sim \mathcal N(0,\sigma_x^2), 
    \qquad
    Y = X_1 + \sigma_{\mathrm{ch}} \varepsilon,
    \quad
    \varepsilon \sim \mathcal N(0,1),
\end{equation}
and a Gaussian smoothing path of the form
\begin{equation}
    X_t = X_1 + \sigma(t)\varepsilon', \qquad t\in[0,1],
\end{equation}
whose marginal variance is
\(
    s^2(t) = \sigma_x^2 + \sigma(t)^2.
\)
As shown in Appendix~\ref{app:proof-prop1}, the probability flow ODE preserving this scalar Gaussian path has a linear velocity field
\begin{equation}
    v_t(x) = \frac{\dot s(t)}{s(t)}\,x.
\end{equation}
Choosing the landing time $t^\star$ such that $\sigma(t^\star)=\sigma_{\mathrm{ch}}$ makes $Y$ distributed as $X_{t^\star}$, such that decoding again corresponds to integrating the probability flow ODE from $t= t^\star$ to $t=1$.

\begin{proposition}[High-SNR performance under a scalar Gaussian model]
\label{prop:scalar-gaussian}
Consider the scalar Gaussian model above and the ideal probability flow ODE with velocity field $v_t(x) = \frac{\dot s(t)}{s(t)}x$, with the channel output interpreted as the landing point $X_{t^\star}=Y$ where $\sigma(t^\star)=\sigma_{\mathrm{ch}}$. Then the resulting LTT decoder is a linear estimator
\begin{equation}
    \widehat X_1^{\mathrm{LTT}} = a_{\mathrm{LTT}} Y,
    \qquad
    a_{\mathrm{LTT}} = \frac{\sigma_x}{\sqrt{\sigma_x^2 + \sigma_{\mathrm{ch}}^2}}.
\end{equation}
The MMSE linear estimator is
\begin{equation}
    \widehat X_1^{\mathrm{MMSE}} = a_{\mathrm{MMSE}} Y,
    \qquad
    a_{\mathrm{MMSE}} = \frac{\sigma_x^2}{\sigma_x^2 + \sigma_{\mathrm{ch}}^2},
\end{equation}
and in the high-SNR regime $\sigma_{\mathrm{ch}}^2 \ll \sigma_x^2$, the excess mean-squared error (MSE) of the LTT decoder satisfies
\begin{equation}
    \mathrm{MSE}_{\mathrm{LTT}} - \mathrm{MSE}_{\mathrm{MMSE}}
    = o \!\big(\sigma_{\mathrm{ch}}^4\big).
\end{equation}
\end{proposition}

Proposition~\ref{prop:scalar-gaussian} shows that, for the scalar Gaussian channels, the induced LTT decoder reduces to a linear estimator with essentially the same structure as the classical MMSE estimator and becomes asymptotically optimal as SNR increases. The fact provides an analytical justification for the Gaussian smoothing path and the associated probability flow ODE design: in Gaussian regimes, the proposed LTT construction is fully consistent with classical estimation theory. Therefore, the scalar analysis serves as a simple yet informative proxy for understanding the robustness observed in our high-dimensional image experiments.

\subsection{Complexity--Distortion Trade-off for the ODE Solver}

We next provide a complexity--distortion characterization for the discretized ODE decoder. Let $f(x,t)=v_{\theta}(x,t)$ denote the learned velocity field, and consider continuous-time ODE
\begin{equation}
    \frac{\mathrm{d}}{\mathrm{d}t} x(t) = f(x(t),t), 
    \qquad t\in[t^\star,1],
\end{equation}
with initial condition $x(t^\star)=y$, where $y$ is the channel output at the landing time $t^\star$. Denote $x_{\mathrm{cont}}(1;y)$ the exact solution at time $t=1$, and $x_{\mathrm{E}}^{(N)}(1;y)$ the numerical solution obtained by Euler method with $N$ uniform steps on $[t^\star,1]$.

\begin{assumption}[Lipschitz and bounded vector field]\label{assump:lipschitz-bounded}
There exist constants $L,B>0$ and a compact set $\mathcal X \subset \mathbb R^d$ such that for all $t \in [t^\star,1]$ and all $x,z \in \mathcal X$,
\begin{equation}
    \|f(x,t) - f(z,t)\| \le L \|x - z\|,
    \qquad
    \|f(x,t)\| \le B,
\end{equation}
and both $x_{\mathrm{cont}}(t;y)$ and $x_{\mathrm{E}}^{(N)}(t;y)$ remain in $\mathcal X$ for $t\in[t^\star,1]$.
\end{assumption}
Under Assumption~\ref{assump:lipschitz-bounded}, the global discretization error of the Euler method,
defined as $\|x_{\mathrm{cont}}(t;y) - x_{\mathrm{E}}^{(N)}(t;y)\|$ over $t\in[t^\star,1]$, has the following bound.

\begin{proposition}[Euler discretization error]
\label{prop:euler-complexity}
Under Assumption~1, there exists a constant $C>0$ depending only on $L,B$ and the horizon $T\triangleq 1-t^\star$ such that for any $N\in\mathbb N$ and any initial state $y\in\mathcal X$,
\begin{equation}
    \big\|x_{\mathrm{cont}}(1;y) - x_{\mathrm{E}}^{(N)}(1;y)\big\|
    \le \frac{C}{N}.
\end{equation}
\end{proposition}
\noindent The proof is provided in Appendix \ref{app:proof-prop2}.
Proposition~\ref{prop:euler-complexity} shows that the discretization error of Euler decoding decays at the order of $1/N$; higher-order solvers only improve this rate, so
Euler method provides a conservative characterization.

We now relate this discretization error to the end-to-end reconstruction error. Let $X_1$ denote the clean image and $Y$ the channel observation. Define the continuous-time and Euler reconstructions as
\begin{equation}
    \widehat X_1^{\mathrm{cont}}(Y) = x_{\mathrm{cont}}(1;Y),
    \qquad
    \widehat X_1^{(N)}(Y) = x_{\mathrm{E}}^{(N)}(1;Y),
\end{equation}
and the corresponding MSEs
\begin{equation}
    \mathrm{MSE}_{\mathrm{cont}}
    = \mathbb E\big[\|X_1 - \widehat X_1^{\mathrm{cont}}(Y)\|^2\big],
\end{equation}
\begin{equation}
     \mathrm{MSE}_{N}
    = \mathbb E\big[\|X_1 - \widehat X_1^{(N)}(Y)\|^2\big].   
\end{equation}

\begin{proposition}[Convergence rate of Euler decoding]
\label{prop:convergence-rate}
Under Assumption~1, 
\begin{equation}
    \mathrm{MSE}_{N} - \mathrm{MSE}_{\mathrm{cont}}
    = \mathcal O\!\big(\tfrac{1}{N}\big),
\end{equation}
the MSE of the discretized decoder converges to that of the continuous-time ODE decoder at rate $1/N$ as the number of ODE steps increases.
\end{proposition}
\noindent The proof is provided in Appendix \ref{app:proof-prop3}.

From a system perspective, each Euler step requires a single evaluation of the neural velocity field $v_{\theta}$. Hence, the decoding complexity scales linearly with the number of steps $N$, while the discretization-induced excess distortion relative to the continuous-time limit decays on the order of $1/N$. This leads to a clear complexity–distortion tradeoff: increasing $N$ incurs a linear increase in computational cost but yields progressively improved reconstruction quality.

\section{Extension to Rayleigh and MIMO channels}
\label{sec:rayleigh-mimo}
In what follows, we will extend our results to Rayleigh fading and MIMO channels.

\subsection{Rayleigh Fading Channels}
Considering a scalar complex Rayleigh fading channel with perfect channel state information at the receiver (CSIR) but not at the transmitter (CSIT), we have,
\begin{equation}
    Y = H\,X_{1} + \sigma_{\mathrm{ch}}\,\varepsilon, 
    \qquad H \sim \mathcal{CN}(0,1),\; \varepsilon \sim \mathcal{CN}(0,1),
    \label{eq:rayleigh-siso}
\end{equation}
where $X_{1}$ denotes the transmitted symbol (a complex entry of the data vector used in Sec.~\ref{sec:land-then-transport}).  
We assume a zero-mean circularly symmetric prior $X_{1} \sim \mathcal{CN}(0,\sigma_x^2)$. 
Given a channel realization $\hat H$, the linear MMSE equalizer that estimates $X_{1}$ from $Y$ is
\begin{equation}
w_{\mathrm{MMSE}}(\hat H)
= \frac{\sigma_x^2\,\hat H^{\ast}}{|\hat H|^2\sigma_x^2 + \sigma_{\mathrm{ch}}^2}
= \frac{\hat H^{\ast}}{|\hat H|^2 + \lambda},
\qquad 
\lambda \triangleq \frac{\sigma_{\mathrm{ch}}^2}{\sigma_x^2}.
\label{eq:mmse-weight}
\end{equation}
Applying \eqref{eq:mmse-weight} yields the pre-equalized observation
\begin{equation}
Z \;\triangleq\; w_{\mathrm{MMSE}}(\hat H)\,Y
= \underbrace{\frac{|\hat H|^2}{|\hat H|^2+\lambda}}_{\alpha(\hat H)\in(0,1)}\,X_{1}
\;+\;
\underbrace{\frac{\sigma_{\mathrm{ch}}\,|\hat H|}{|\hat H|^2+\lambda}}_{\sigma_{\mathrm{eff}}(\hat H)}\,\varepsilon',
\label{eq:zw-awgn}
\end{equation}
where $\varepsilon' \sim \mathcal{CN}(0,1)$. Thus, conditioned on $\hat H$, the random variable $Z$ is an \emph{AWGN} observation of $X_{1}$ with the mean multiplied by 
$\alpha(\hat H)$ and an effective noise variance
\begin{equation}
\sigma_{\mathrm{eff}}(\hat H) 
= \frac{|\hat H|}{|\hat H|^2+\lambda}\,\sigma_{\mathrm{ch}}.
\label{eq:sigma-eff}
\end{equation}

Then, we can choose the landing time on the AWGN flow path to match the noise level:
\begin{equation}
t^{\star}(\hat H) \;=\; \sigma^{-1}\!\big(\,\sigma_{\mathrm{eff}}(\hat H)\,\big).
\label{eq:rayleigh-landing}
\end{equation}
With landing time $t^{\star}(\hat H)$, we set the initial condition for the backward ODE to
\begin{equation}
    X_{t^{\star}(\hat H)} = Z.
\end{equation}
Thus, Rayleigh fading channels with linear MMSE equalization are transformed to land the observation on the equivalent AWGN flow path $X_t$ of Sec.~\ref{sec:land-then-transport}.
\subsection{MIMO Channels}

We next consider an $N_t \times N_r$ MIMO channel with perfect CSIR and without CSIT, i.e., no precoding at the transmitter:
\begin{equation}
\mathbf Y = \mathbf H\,\mathbf X_{1} + \sigma_{\mathrm{ch}}\,\boldsymbol\varepsilon, 
\qquad 
\boldsymbol\varepsilon \sim \mathcal{CN}(\mathbf 0,\mathbf I),
\label{eq:mimo-model}
\end{equation}
where $\mathbf H \in \mathbb C^{N_r \times N_t}$ is known at the receiver only, and $\mathbf X_{1}$ denotes the transmitted vector.

Let the receiver-side SVD of $\mathbf H$ (from CSIR) be
\begin{equation}
\mathbf H=\mathbf U\,\boldsymbol\Sigma\,\mathbf V^{\mathrm H}, 
\qquad
\boldsymbol\Sigma=\operatorname{diag}(\sigma_1,\dots,\sigma_r),\; r=\operatorname{rank}(\mathbf H).
\label{eq:mimo-svd}
\end{equation}
Left-rotating the received vector by $\mathbf U^{\mathrm H}$ yields
\begin{equation}
\tilde{\mathbf Y} \triangleq \mathbf U^{\mathrm H}\mathbf Y, 
\qquad
\tilde{\boldsymbol\varepsilon} \triangleq \mathbf U^{\mathrm H}\boldsymbol\varepsilon .
\label{eq:mimo-left-rotate}
\end{equation}
For notational convenience, we represent the (unknown) transmit vector in the right-singular basis as
\begin{equation}
\tilde{\mathbf X}_{1} \triangleq \mathbf V^{\mathrm H}\mathbf X_{1}.
\label{eq:mimo-x-tilde}
\end{equation}
Note that \eqref{eq:mimo-x-tilde} is only a change of coordinates used for receiver-side estimation; it does not imply any transmitter-side multiplication by $\mathbf V$ and hence does not require CSIT. Substituting \eqref{eq:mimo-svd}--\eqref{eq:mimo-x-tilde} into \eqref{eq:mimo-model} gives the SVD domain observation
\begin{equation}
\tilde{\mathbf Y} = \boldsymbol\Sigma\,\tilde{\mathbf X}_{1} + \sigma_{\mathrm{ch}}\,\tilde{\boldsymbol\varepsilon},
\label{eq:svd-parallel}
\end{equation}
i.e., $r$ parallel scalar subchannels
\begin{equation}
\tilde Y_i = \sigma_i\,\tilde X_{1,i} + \sigma_{\mathrm{ch}}\,\tilde\varepsilon_i, 
\qquad i=1,\dots,r.
\label{eq:svd-parallel-scalar}
\end{equation}
Assuming a zero-mean circularly symmetric prior $\tilde X_{1,i} \sim \mathcal{CN}(0,\sigma_x^2)$, the per-mode linear MMSE weight is
\begin{equation}
w_i^{\mathrm{MMSE}}
=\frac{\sigma_x^2\,\sigma_i}{\sigma_i^2\sigma_x^2+\sigma_{\mathrm{ch}}^2}
=\frac{\sigma_i}{\sigma_i^2+\lambda},
\qquad \lambda \triangleq \frac{\sigma_{\mathrm{ch}}^2}{\sigma_x^2},
\label{eq:mimo-mmse-w}
\end{equation}
which is consistent with single-input single-output (SISO) Rayleigh fading channels. Applying $w_i^{\mathrm{MMSE}}$ to \eqref{eq:svd-parallel} gives
\begin{equation}
\hat Z_i \triangleq w_i^{\mathrm{MMSE}}\tilde Y_i
= \underbrace{\frac{\sigma_i^2}{\sigma_i^2+\lambda}}_{\alpha_i\in(0,1)}\,\tilde X_{1,i}
\;+\; 
\underbrace{\frac{\sigma_{\mathrm{ch}}\sigma_i}{\sigma_i^2+\lambda}}_{\sigma_{\mathrm{eff},i}}\,\tilde\varepsilon_i,
\label{eq:mimo-awgn-like}
\end{equation}
i.e., an \emph{AWGN-equivalent} $\tilde X_{1,i}$ with a mean modified factor $\alpha_i$ and effective noise variance
\begin{equation}
\sigma_{\mathrm{eff},i}=\frac{\sigma_{\mathrm{ch}}\sigma_i}{\sigma_i^2+\lambda}.
\label{eq:mimo-sigma-eff}
\end{equation}
Therefore, for the $i$-th subchannel, the landing time $t_i^\star$ on the AWGN flow path is determined by matching the effective noise level:
\begin{equation}
t_i^{\star} = \sigma^{-1}(\sigma_{\mathrm{eff}, i})
= \sigma^{-1}\!\left(\frac{\sigma_{\mathrm{ch}}\sigma_i}{\sigma_i^2+\lambda}\right),
\qquad i=1,\dots,r.
\end{equation}
For decoding at the receiver, with landing time $t_i^{\star}$, we set the initial condition of the backward ODE on each subchannel to 
\begin{equation}
X_{t_i^{\star},\, i} = \hat Z_i, \qquad i=1,\dots,r.
\end{equation}
After integrating the ODE from $t= t_i^{\star}$ to $1$ for all modes using the same learned velocity field $v_{\theta}$, we obtain $\widehat{\tilde{\mathbf X}}_{1}$ and rotate back via
\begin{equation}
\widehat{\mathbf X}_{1} = \mathbf V\,\widehat{\tilde{\mathbf X}}_{1},
\end{equation}
yielding a MIMO decoder that reuses the AWGN flow path and student field trained in Sec.~\ref{sec:land-then-transport}.


\subsection{Training and decoding in Rayleigh and MIMO channels}

\begin{figure*}[t]
    \centering 
    \subfloat[AWGN - PSNR]{\includegraphics[width=0.24\textwidth]{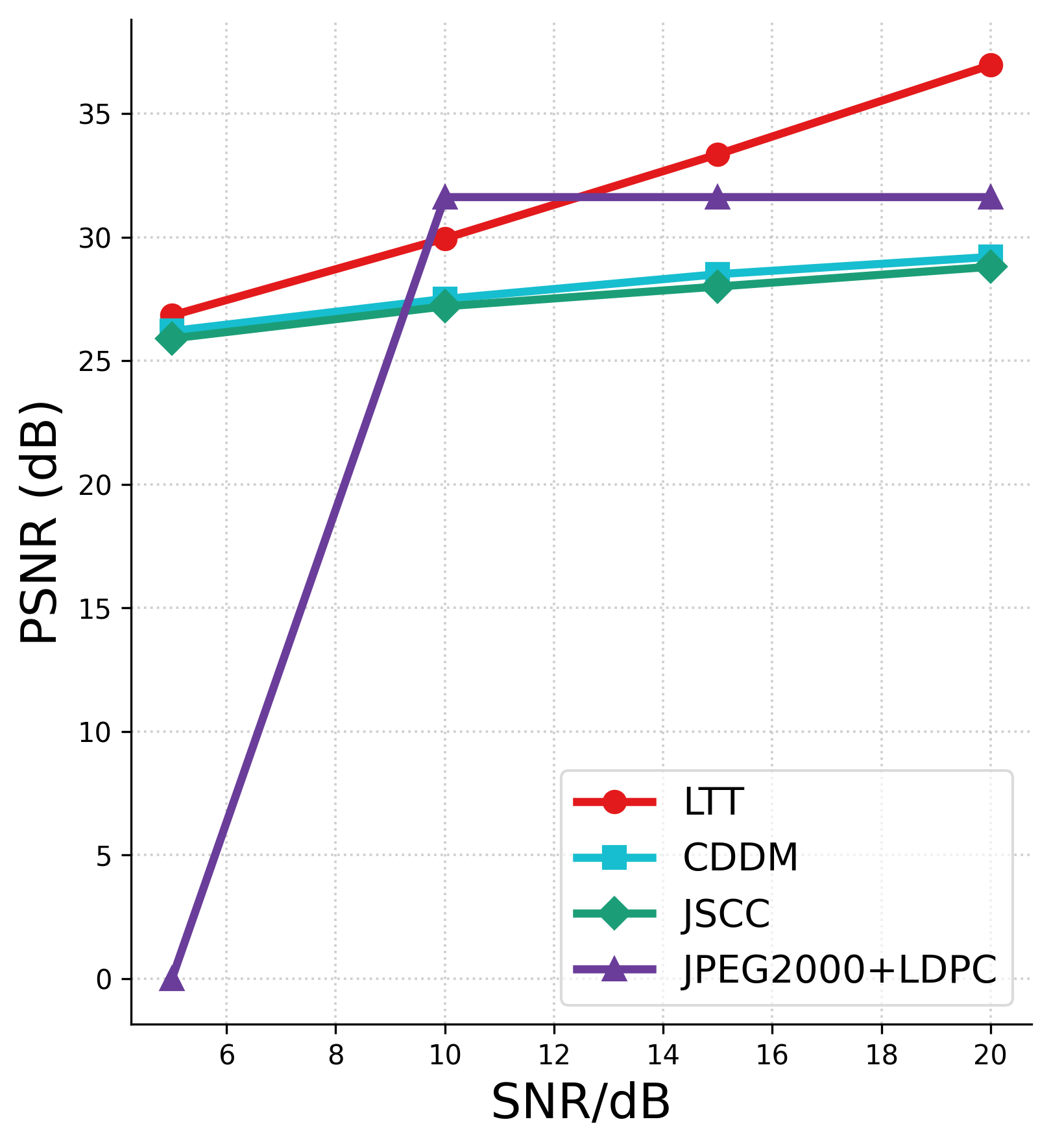}
        \label{fig:2baseling1}}
    \hfill 
    \subfloat[AWGN - MS-SSIM]{\includegraphics[width=0.24\textwidth]{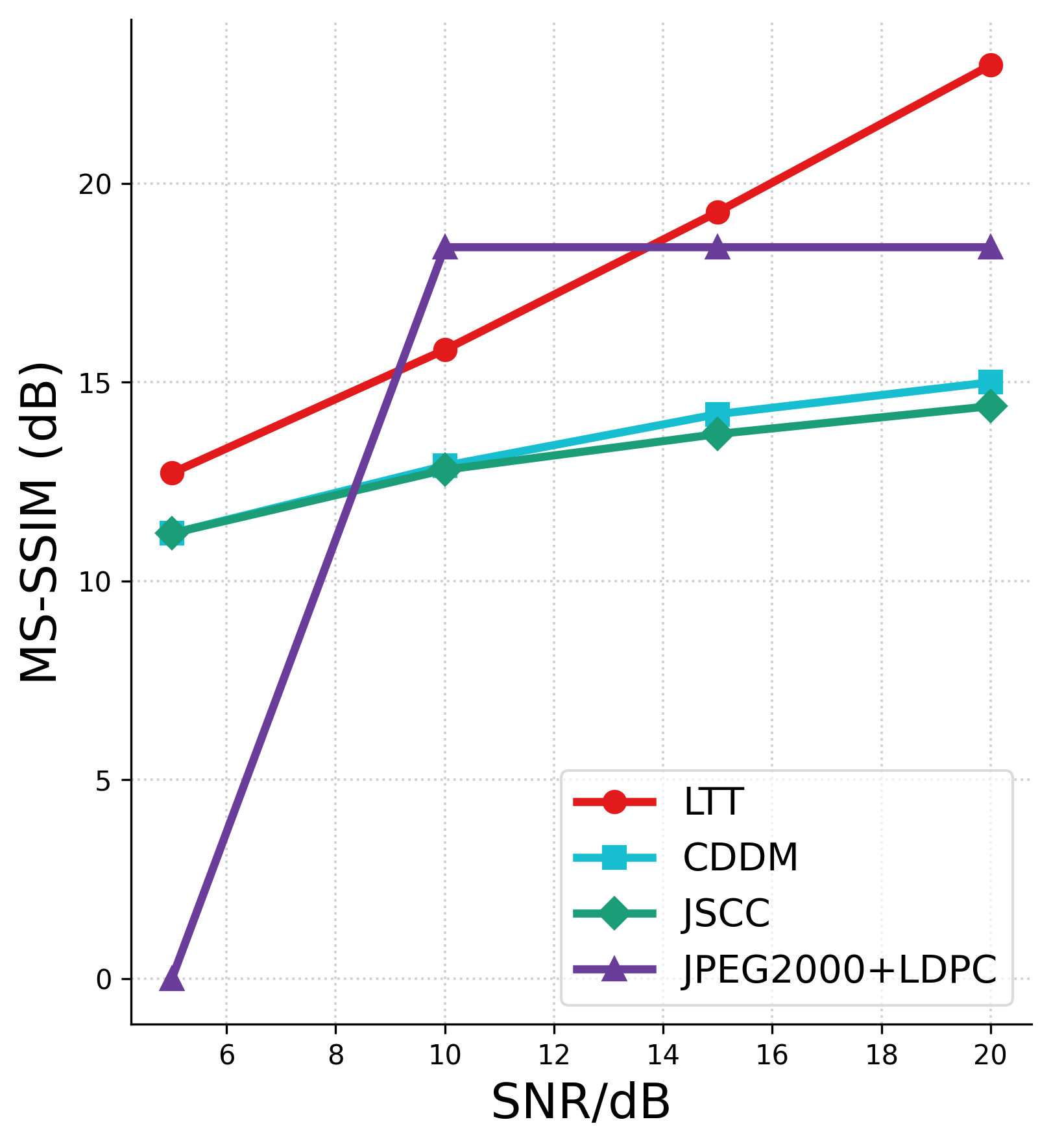}
        \label{fig:2baseling2}}
    \hfill 
    \subfloat[Rayleigh - PSNR]{\includegraphics[width=0.24\textwidth]{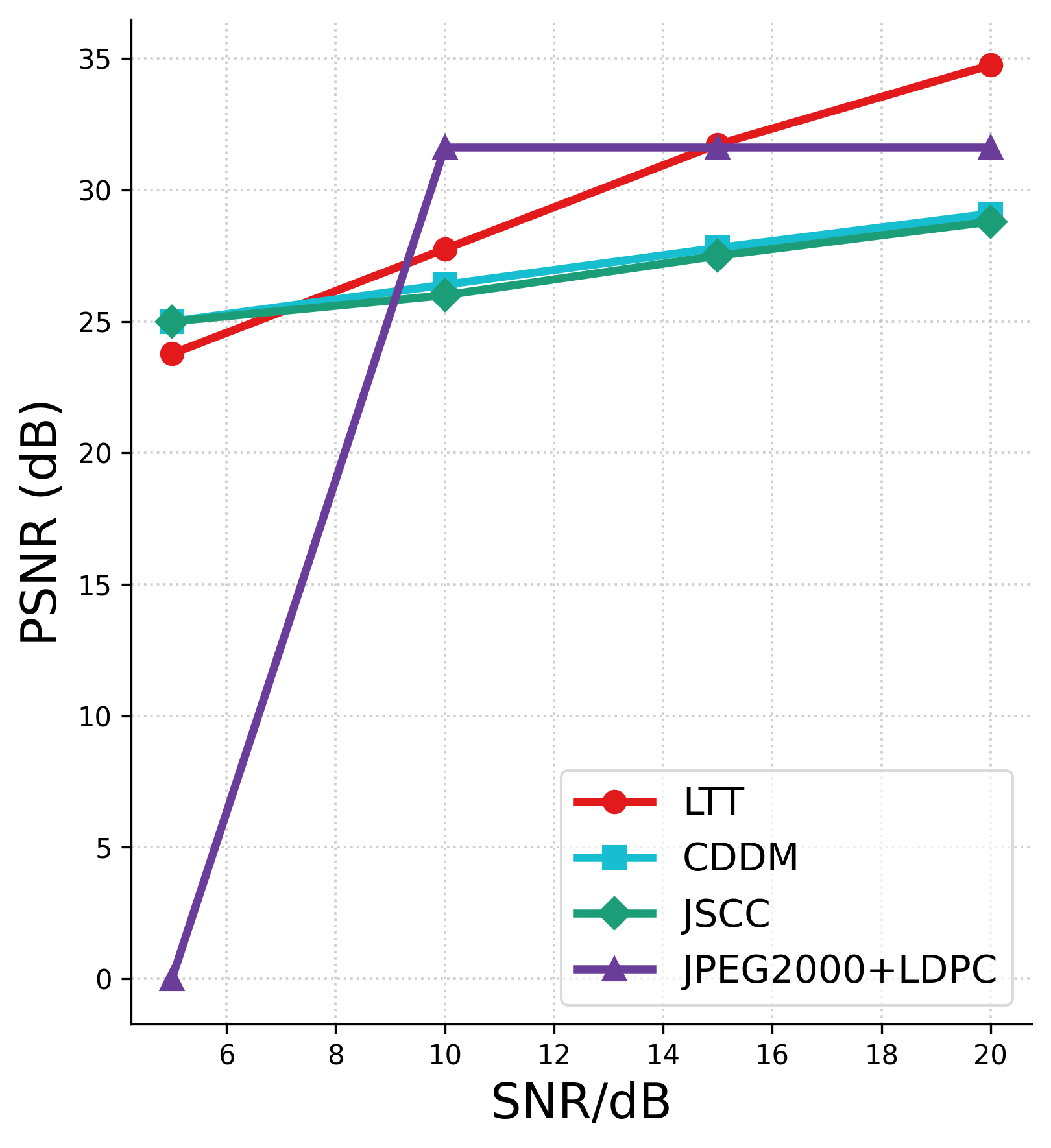}
        \label{fig:2baseling3}}
    \hfill
    \subfloat[Rayleigh - MS-SSIM]{\includegraphics[width=0.24\textwidth]{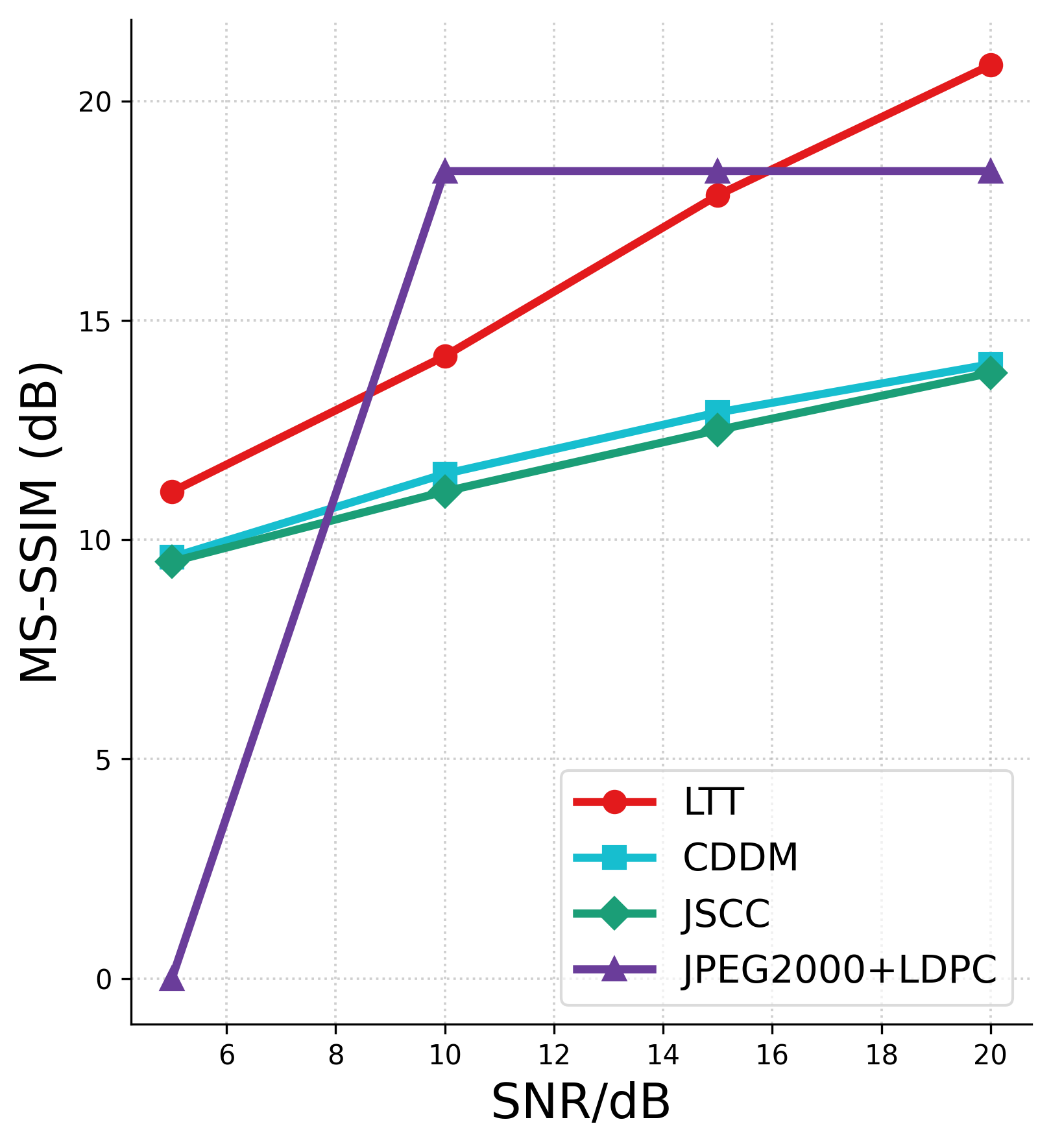}
        \label{fig:2baseling4}}
    \caption{Performance compared to baseline models in AWGN and Rayleigh channels on DIV2K dataset.}
    \label{fig:four_images}
\end{figure*}

The results above imply that extending the proposed decoder from AWGN to Rayleigh and MIMO channels requires no additional training. 
The linear MMSE front-ends in~\eqref{eq:zw-awgn} and~\eqref{eq:mimo-awgn-like} convert each channel realization into an AWGN-equivalent channel with effective noise level $\sigma_{\mathrm{eff}}$ (or $\sigma_{\mathrm{eff},i}$), which in turn defines a landing time $t^\star$ (or $t_i^\star$) along the original AWGN path.
The procedure can be summarized as follows:
\begin{itemize}
    \item \emph{Training phase:} Train $v_{\theta}$ once under the AWGN channel assumption by sampling noisy pairs $(x_1, X_t)$ according to $\sigma(t)$, as described in Section~\ref{sec:proposed}.  
    \item \emph{Decoding phase:} For each channel use, we apply the corresponding linear MMSE equalizer to obtain $z$ (Rayleigh) or $\hat z_i$ (MIMO), compute the effective noise level and landing time via $\sigma_{\mathrm{eff}} \mapsto t^\star = \sigma^{-1}(\sigma_{\mathrm{eff}})$ or $\sigma_{\mathrm{eff},i}\mapsto t^\star = \sigma^{-1}(\sigma_{\mathrm{eff,i}})$, and then integrate the same ODE driven by $v_{\theta}(x,t)$ from $t^\star$ to $t=1$.
\end{itemize}

In such way, the physical channel and the linear front-end perform the \emph{landing step} by mapping the observation to an AWGN-equivalent sample at time $t^\star$ on the same probability path, and the learned ODE performs the \emph{transport step} by deterministically evolving the sample from $t^\star$ to $t=1$ to obtain the estimate of clean images. Therefore, for any linear Gaussian channels (e.g., Rayleigh fading and MIMO channels) that admit an AWGN-equivalent representation with effective noise variance, the decoder trained for AWGN channels remains applicable. 



\section{Numerical Results}
\label{sec:results}

\begin{table*}[t]
    \centering
    \caption{LTT model performance over AWGN, Rayleigh, and MIMO channels on DIV2K dataset.}
    \label{tab:awgn}
    \begin{tabular}{
        c 
        S[table-format=2.3] S[table-format=2.3] S[table-format=1.4] S[table-format=2.3] 
        S[table-format=2.3] S[table-format=2.3] S[table-format=1.4] S[table-format=1.3] 
        S[table-format=2.3] S[table-format=2.3] S[table-format=1.4] S[table-format=1.3]  
    }
        \toprule
        \multirow{3}{*}{SNR (dB)}
        & \multicolumn{4}{c}{AWGN}
        & \multicolumn{4}{c}{Rayleigh}
        & \multicolumn{4}{c}{MIMO} \\
        \cmidrule(lr){2-5} \cmidrule(lr){6-9} \cmidrule(lr){10-13}
        & \multicolumn{1}{c}{PSNR}
        & \multicolumn{1}{c}{MS-SSIM}
        & \multicolumn{1}{c}{LPIPS}
        & \multicolumn{1}{c}{$\Delta$PSNR}
        & \multicolumn{1}{c}{PSNR}
        & \multicolumn{1}{c}{MS-SSIM}
        & \multicolumn{1}{c}{LPIPS}
        & \multicolumn{1}{c}{$\Delta$PSNR}
        & \multicolumn{1}{c}{PSNR}
        & \multicolumn{1}{c}{MS-SSIM}
        & \multicolumn{1}{c}{LPIPS}
        & \multicolumn{1}{c}{$\Delta$PSNR} \\
        & \multicolumn{1}{c}{(dB) $\uparrow$}
        & \multicolumn{1}{c}{(dB) $\uparrow$}
        & \multicolumn{1}{c}{$\downarrow$}
        & \multicolumn{1}{c}{(dB) $\uparrow$}
        & \multicolumn{1}{c}{(dB) $\uparrow$}
        & \multicolumn{1}{c}{(dB) $\uparrow$}
        & \multicolumn{1}{c}{$\downarrow$}
        & \multicolumn{1}{c}{(dB) $\uparrow$}
        & \multicolumn{1}{c}{(dB) $\uparrow$}
        & \multicolumn{1}{c}{(dB) $\uparrow$}
        & \multicolumn{1}{c}{$\downarrow$}
        & \multicolumn{1}{c}{(dB) $\uparrow$} \\
        \midrule
        0  & 24.830 & 10.950 & 0.3860 & 12.002
           & 19.675 &  7.914 & 0.4758 &  6.528
           & 20.217 &  8.978 & 0.3965 &  4.347 \\
        3  & 26.466 & 12.650 & 0.3225 & 11.094
           & 21.971 &  9.588 & 0.4237 &  7.378
           & 22.782 & 10.841 & 0.3409 &  5.374 \\
        5  & 27.596 & 13.844 & 0.2803 & 10.437
           & 23.888 & 11.080 & 0.3742 &  7.608
           & 25.401 & 12.761 & 0.2908 &  6.155 \\
        7  & 28.774 & 15.052 & 0.2380 &  9.772
           & 24.753 & 11.771 & 0.3524 &  7.529
           & 26.167 & 13.503 & 0.2677 &  6.314 \\
        10 & 30.603 & 16.940 & 0.1811 &  8.786
           & 27.746 & 14.335 & 0.2715 &  7.582
           & 28.014 & 15.135 & 0.2269 &  6.109 \\
        12 & 31.869 & 18.212 & 0.1476 &  8.137
           & 29.144 & 15.419 & 0.2388 &  7.467
           & 31.841 & 18.270 & 0.1540 &  6.433 \\
        15 & 33.829 & 20.201 & 0.1055 &  7.198
           & 31.920 & 18.165 & 0.1752 &  6.774
           & 33.412 & 19.910 & 0.1246 &  6.161 \\
        \bottomrule
    \end{tabular}
\end{table*}

\begin{table*}[t]
    \centering
    \caption{LTT model performance over AWGN, Rayleigh, and MIMO channels on MNIST dataset.}
    \label{tab:rayleigh}
    \begin{tabular}{
        c 
        S[table-format=2.3] S[table-format=2.3] S[table-format=1.4] S[table-format=2.3] 
        S[table-format=2.3] S[table-format=2.3] S[table-format=1.4] S[table-format=1.3] 
        S[table-format=2.3] S[table-format=2.3] S[table-format=1.4] S[table-format=1.3]  
    }
        \toprule
        \multirow{3}{*}{SNR (dB)}
        & \multicolumn{4}{c}{AWGN}
        & \multicolumn{4}{c}{Rayleigh}
        & \multicolumn{4}{c}{MIMO} \\
        \cmidrule(lr){2-5} \cmidrule(lr){6-9} \cmidrule(lr){10-13}
        & \multicolumn{1}{c}{PSNR}
        & \multicolumn{1}{c}{MS-SSIM}
        & \multicolumn{1}{c}{LPIPS}
        & \multicolumn{1}{c}{$\Delta$PSNR}
        & \multicolumn{1}{c}{PSNR}
        & \multicolumn{1}{c}{MS-SSIM}
        & \multicolumn{1}{c}{LPIPS}
        & \multicolumn{1}{c}{$\Delta$PSNR}
        & \multicolumn{1}{c}{PSNR}
        & \multicolumn{1}{c}{MS-SSIM}
        & \multicolumn{1}{c}{LPIPS}
        & \multicolumn{1}{c}{$\Delta$PSNR} \\
        & \multicolumn{1}{c}{(dB) $\uparrow$}
        & \multicolumn{1}{c}{(dB) $\uparrow$}
        & \multicolumn{1}{c}{$\downarrow$}
        & \multicolumn{1}{c}{(dB) $\uparrow$}
        & \multicolumn{1}{c}{(dB) $\uparrow$}
        & \multicolumn{1}{c}{(dB) $\uparrow$}
        & \multicolumn{1}{c}{$\downarrow$}
        & \multicolumn{1}{c}{(dB) $\uparrow$}
        & \multicolumn{1}{c}{(dB) $\uparrow$}
        & \multicolumn{1}{c}{(dB) $\uparrow$}
        & \multicolumn{1}{c}{$\downarrow$}
        & \multicolumn{1}{c}{(dB) $\uparrow$}\\
        \midrule
        0  & 20.460 &  6.155 & 0.1702 &  9.946
           & 12.426 &  2.799 & 0.4816 &  3.710
           & 14.883 &  4.115 & 0.3016 &  4.187 \\
        3  & 22.148 &  7.266 & 0.1415 &  9.150
           & 14.338 &  3.980 & 0.3649 &  4.407
           & 16.715 &  5.225 & 0.2178 &  4.582 \\
        5  & 23.321 &  8.047 & 0.1240 &  8.544
           & 15.725 &  4.891 & 0.2943 &  4.760
           & 18.115 &  6.081 & 0.1689 &  4.820 \\
        7  & 24.541 &  8.913 & 0.1070 &  7.936
           & 16.816 &  5.661 & 0.2476 &  4.857
           & 19.567 &  6.929 & 0.1327 &  5.025 \\
        10 & 26.428 & 10.273 & 0.0850 &  7.024
           & 18.707 &  6.947 & 0.1870 &  4.874
           & 21.422 &  8.049 & 0.0981 &  5.063 \\
        12 & 27.744 & 11.227 & 0.0715 &  6.436
           & 19.881 &  7.785 & 0.1573 &  4.670
           & 23.052 &  8.887 & 0.0786 &  5.105 \\
        15 & 29.810 & 12.778 & 0.0541 &  5.605
           & 21.874 &  9.112 & 0.1215 &  4.385
           & 25.279 & 10.142 & 0.0611 &  5.015 \\
        \bottomrule
    \end{tabular}
\end{table*}

\begin{table*}[t]
    \centering
    \caption{LTT model performance over AWGN, Rayleigh, and MIMO channels on Fashion-MNIST dataset.}
    \label{tab:mimo}
    \begin{tabular}{
        c 
        S[table-format=2.3] S[table-format=2.3] S[table-format=1.4] S[table-format=2.3] 
        S[table-format=2.3] S[table-format=2.3] S[table-format=1.4] S[table-format=1.3] 
        S[table-format=2.3] S[table-format=2.3] S[table-format=1.4] S[table-format=1.3]  
    }
        \toprule
        \multirow{3}{*}{SNR (dB)}
        & \multicolumn{4}{c}{AWGN}
        & \multicolumn{4}{c}{Rayleigh}
        & \multicolumn{4}{c}{MIMO} \\
        \cmidrule(lr){2-5} \cmidrule(lr){6-9} \cmidrule(lr){10-13}
        & \multicolumn{1}{c}{PSNR}
        & \multicolumn{1}{c}{MS-SSIM}
        & \multicolumn{1}{c}{LPIPS}
        & \multicolumn{1}{c}{$\Delta$PSNR}
        & \multicolumn{1}{c}{PSNR}
        & \multicolumn{1}{c}{MS-SSIM}
        & \multicolumn{1}{c}{LPIPS}
        & \multicolumn{1}{c}{$\Delta$PSNR}
        & \multicolumn{1}{c}{PSNR}
        & \multicolumn{1}{c}{MS-SSIM}
        & \multicolumn{1}{c}{LPIPS}
        & \multicolumn{1}{c}{$\Delta$PSNR} \\
        & \multicolumn{1}{c}{(dB) $\uparrow$}
        & \multicolumn{1}{c}{(dB) $\uparrow$}
        & \multicolumn{1}{c}{$\downarrow$}
        & \multicolumn{1}{c}{(dB) $\uparrow$}
        & \multicolumn{1}{c}{(dB) $\uparrow$}
        & \multicolumn{1}{c}{(dB) $\uparrow$}
        & \multicolumn{1}{c}{$\downarrow$}
        & \multicolumn{1}{c}{(dB) $\uparrow$}
        & \multicolumn{1}{c}{(dB) $\uparrow$}
        & \multicolumn{1}{c}{(dB) $\uparrow$}
        & \multicolumn{1}{c}{$\downarrow$}
        & \multicolumn{1}{c}{(dB) $\uparrow$}\\
        \midrule
        0  & 19.827 &  4.976 & 0.2813 &  9.304
           & 12.051 &  2.289 & 0.4804 &  2.037
           & 13.792 &  3.544 & 0.3892 &  1.537 \\
        3  & 21.655 &  6.154 & 0.2324 &  8.646
           & 13.856 &  3.314 & 0.3909 &  2.634
           & 15.610 &  4.641 & 0.3160 &  1.926 \\
        5  & 22.899 &  7.004 & 0.2016 &  8.109
           & 15.285 &  4.159 & 0.3268 &  3.036
           & 17.040 &  5.495 & 0.2649 &  2.224 \\
        7  & 24.153 &  7.901 & 0.1732 &  7.538
           & 16.507 &  4.910 & 0.2786 &  3.273
           & 18.574 &  6.432 & 0.2149 &  2.540 \\
        10 & 26.102 &  9.376 & 0.1334 &  6.687
           & 18.625 &  6.240 & 0.2146 &  3.558
           & 20.715 &  7.753 & 0.1589 &  2.910 \\
        12 & 27.460 & 10.447 & 0.1091 &  6.143
           & 19.943 &  7.115 & 0.1811 &  3.550
           & 22.506 &  8.813 & 0.1236 &  3.171 \\
        15 & 29.557 & 12.086 & 0.0779 &  5.341
           & 22.132 &  8.454 & 0.1427 &  3.503
           & 25.038 & 10.358 & 0.0901 &  3.441 \\
        \bottomrule
    \end{tabular}
\end{table*}

\subsection{Experimental Setups}
\subsubsection{Datasets}
We evaluate the proposed methods with three common image datasets: MNIST~\cite{lecun1998gradient}, Fashion-MNIST~\cite{xiao2017fashionmnist}, and DIV2K~\cite{agustsson2017ntire}. MNIST and Fashion-MNIST contain $60{,}000$ training and $10{,}000$ test gray-scale images of $28\times 28$ handwritten digits and clothing objectives, respectively, which are used as low-resolution examples. DIV2K comprises $800$ training and $100$ validation natural images; all images are cropped and resized to $256\times 256$ before use. For each dataset, $10\%$ of the training images are separated for validation, and we report results on the standard test set (MNIST/Fashion-MNIST) or official validation set (DIV2K).

\subsubsection{Baselines}
We compare the proposed decoder with three common baselines:
\begin{itemize}
    \item JPEG2000+LDPC~\cite{ETSI_EN_302_755}: A separated source–channel coding baseline using JPEG2000 followed by DVB-S2 LDPC (block length 64\,800, rate $1/2$). We use a compression ratio of 16 for AWGN and 8 for Rayleigh and MIMO channels.
    \item DeepJSCC~\cite{bourtsoulatze2019deep}: A DNN-based JSCC scheme mapping images directly to channel symbols and reconstructing from noisy observations. The number of transmitted symbols is matched to that of our method.
    \item CDDM~\cite{cddm}: A diffusion-based generative decoder that applies a score-based diffusion model at the receiver to refine reconstructions from noisy channel outputs.
\end{itemize}

\subsubsection{Performance metrics}  We evaluate reconstruction quality using four metrics: PSNR, MS-SSIM~\cite{wang2003multiscale}, learned perceptual image patch similarity (LPIPS)~\cite{zhang2018unreasonable}, and $\Delta\text{PSNR}$.  PSNR measures pixel-wise fidelity. MS-SSIM captures multi-scale structural similarity, and LPIPS quantifies perceptual distance in a deep feature space, where lower values indicate better quality. $\Delta\text{PSNR}$ denotes the PSNR gain over the directly received noisy image, i.e., the difference between the PSNR of the reconstructed image and the noisy channel output.

\subsubsection{Implementation}
All models are implemented in PyTorch and trained on a single NVIDIA A40 GPU. The maximum noise level of the smoothing path is set to $\sigma_{\max} = 1.0$, which corresponds to an effective SNR range covering above $0$dB in our experiments. Unless otherwise stated, we train for $50$ epochs with a learning rate of $1 \times 10^{-3}$. The number of ODE steps is set to 10. The batch size is set to $64$ for MNIST and Fashion-MNIST, and $32$ for DIV2K. For MIMO channels, we simulate a $2 \times 2$ MIMO system.

\subsection{Result Analysis}

\begin{figure*}[htbp]
    \centering
    \subfloat[Original]{%
        \includegraphics[width=0.22\textwidth]{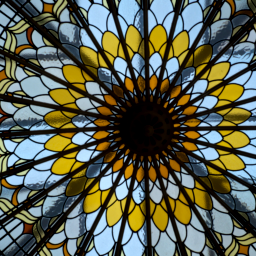}
    }\hfill
    \subfloat[DeepJSCC]{%
        \includegraphics[width=0.22\textwidth]{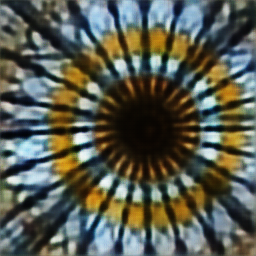}
    }\hfill
    \subfloat[JPEG+LDPC]{%
        \includegraphics[width=0.22\textwidth]{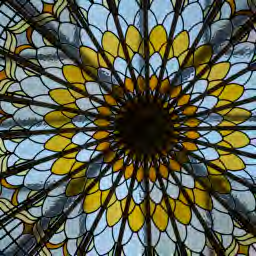}
    }\hfill
    \subfloat[LTT]{%
        \includegraphics[width=0.22\textwidth]{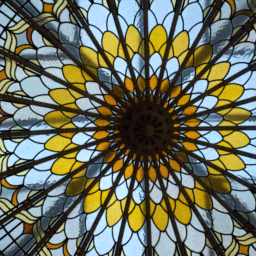}
    }\\[2mm]

    \subfloat[Original]{%
        \includegraphics[width=0.22\textwidth]{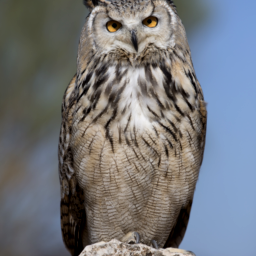}
    }\hfill
    \subfloat[DeepJSCC]{%
        \includegraphics[width=0.22\textwidth]{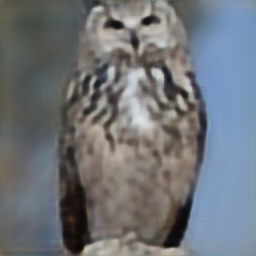}
    }\hfill
    \subfloat[JPEG+LDPC]{%
        \includegraphics[width=0.22\textwidth]{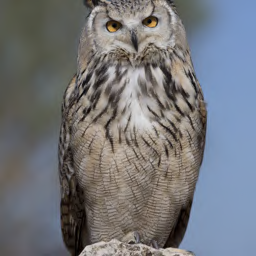}
    }\hfill
    \subfloat[LTT]{%
        \includegraphics[width=0.22\textwidth]{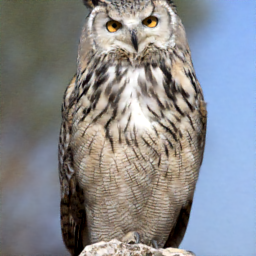}
    }\\[2mm]

    \subfloat[Original]{%
        \includegraphics[width=0.22\textwidth]{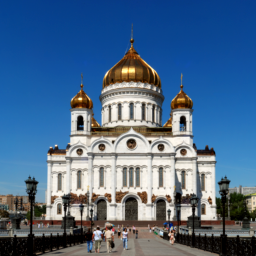}
    }\hfill
    \subfloat[DeepJSCC]{%
        \includegraphics[width=0.22\textwidth]{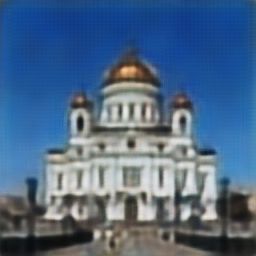}
    }\hfill
    \subfloat[JPEG+LDPC]{%
        \includegraphics[width=0.22\textwidth]{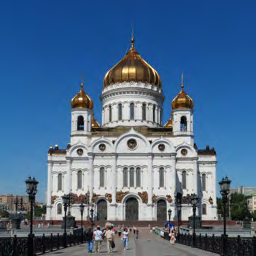}
    }\hfill
    \subfloat[LTT]{%
        \includegraphics[width=0.22\textwidth]{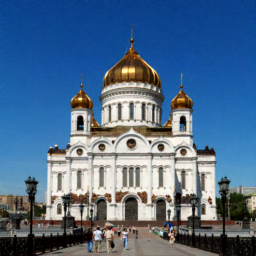}
    }

    \caption{The visual comparison of reconstructed images on the DIV2K dataset at 20\,dB. The first row shows results over AWGN channel, the second row over Rayleigh fading channel, and the third row over MIMO channel. For each channel condition, our method is compared to DeepJSCC and JPEG+LDPC methods.}
\label{visual}
\end{figure*}

\subsubsection{Performance compared with baseline models}
Fig.~\ref{fig:four_images} compares the proposed decoder with CDDM, DeepJSCC, and JPEG2000+LDPC on DIV2K with both AWGN and Rayleigh fading channels. Compared with CDDM and DeepJSCC, the proposed decoder consistently provides higher reconstruction quality. In AWGN channel at $\text{SNR}=20$~dB, our decoder improves PSNR by $26.6\%$ and $28.3\%$ over CDDM and DeepJSCC, respectively, and increases MS-SSIM by $53.2\%$ and $59.6\%$. In Rayleigh fading channels at the same transmitting power, the PSNR gains reach $19.4\%$ and $20.7\%$, while the MS-SSIM gains are $48.7\%$ and $50.8\%$. Similar trends hold at lower SNRs, where the proposed decoder maintains competitive PSNR and consistently higher MS-SSIM, indicating more faithful perceptual reconstruction than CDDM and DeepJSCC baselines. For JPEG2000+LDPC, we fix the end-to-end bandwidth efficiency (i.e., the number of channel uses per source pixel) to be identical to that of our scheme for a fair comparison. Under this setting, JPEG2000+LDPC exhibits a pronounced cliff effect: at low SNRs, JPEG2000+LDPC systems frequently fail to decode and the reconstruction quality drops to nearly zero, whereas once the SNR exceeds the decoding threshold it can deliver high PSNR and MS-SSIM. However, JPEG2000+LDPC performance quickly saturates and shows almost no further improvement when SNR increases. In contrast, the proposed flow-based decoder degrades gracefully in the low-SNR regime and continues to benefit from better channel conditions. At $\text{SNR}=20$~dB, our decoder achieves PSNR/MS-SSIM gains of $16.9\%/24.9\%$ in AWGN channels and $9.9\%/13.1\%$ in Rayleigh channels over JPEG2000+LDPC channels, demonstrating superior rate–distortion performance across a wide SNR range.

\subsubsection{Model performance}
Tables~\ref{tab:awgn},~\ref{tab:rayleigh} and~\ref{tab:mimo} summarize the quantitative performance of the proposed decoder over AWGN, Rayleigh, and MIMO channels on DIV2K, MNIST, and Fashion-MNIST datasets, respectively. For DIV2K, proposed LTT decoder achieves up to 33.83~dB, 31.92~dB, and 33.41~dB PSNR at 15~dB SNR under AWGN, Rayleigh, and MIMO channels respectively, with the corresponding MS-SSIM exceeding 20~dB for AWGN/MIMO and LPIPS reduced below $0.11$.  $\Delta$PSNR column shows large gains over the best baseline, ranging from about 7--12~dB on AWGN, 6--8~dB on Rayleigh, and 4--6~dB on MIMO channels across the considered SNRs. Similar trends are observed on MNIST and Fashion-MNIST: for AWGN channels, our method reaches around 30~dB PSNR with LPIPS close to $0.05$, while maintaining consistent improvements of approximately 5--10~dB in $\Delta$PSNR; under Rayleigh and MIMO channels, our method still provides 2--5~dB average PSNR gains together with higher MS-SSIM and lower LPIPS.  Simulations for MIMO channels consistently outperform those of Rayleigh channels due to spatial diversity and array gain. Overall, these results show that an  LTT decoder trained for AWGN channels, can  also be used in Rayleigh and MIMO channels via MMSE-based equalization. The proposed methods show robustness across datasets and channel models.

\subsubsection{Visualization}
Fig.~\ref{visual} provides visual comparisons across three channel conditions on DIV2K dataset at 20\,dB. DeepJSCC consistently produces reconstructions with severe loss of fine textures, while JPEG2000+LDPC preserves more structure but introduces noticeable compression artifacts and color inconsistencies, especially under fading and MIMO channels. In contrary, our method yields sharper edges, cleaner textures, and more faithful geometric details across all examples, demonstrating its robustness to channel distortion and clear advantage in perceptual reconstruction quality.

\subsubsection{Ablation on ODE steps}

Table~\ref{tab:ode_steps_10dB} shows that increasing ODE steps lead to only minor performance variations at 10 dB: PSNR stays within 30.10–30.52 dB and MS-SSIM within 16.53–16.83 dB, while LPIPS also fluctuates in a narrow range without a clear monotonic trend. In contrast, the per-sample latency grows almost linearly from 0.18s to 1.80s, implying a $10\times$ increase in computational cost for negligible quality gains. Balancing accuracy and efficiency, we use a 10-step configuration in all experiments, as it provides the highest reconstruction quality under a moderate computational budget. Compared with diffusion-based decoders typically requiring tens to hundreds of stochastic denoising steps, our ODE-based decoder achieves competitive or better reconstruction quality with as few as 10 deterministic steps, leading to significantly reduced decoding latency. The result is consistent with the complexity–distortion tradeoff characterized in Proposition~\ref{prop:euler-complexity}, where the reconstruction error approaches the continuous-time limit as ODE steps increases.

\begin{table}[t]
    \centering
    \caption{Ablation study on the number of ODE steps for reconstruction quality and per-sample inference time at 10\,dB SNR on DIV2K dataset.}
    \label{tab:ode_steps_10dB}
    \begin{tabular}{
        S[table-format=2.0]   
        S[table-format=2.3]   
        S[table-format=2.3]   
        S[table-format=1.4]   
        S[table-format=1.4]   
    }
        \toprule
        \multicolumn{1}{c}{Steps}
        & \multicolumn{1}{c}{PSNR}
        & \multicolumn{1}{c}{MS-SSIM}
        & \multicolumn{1}{c}{LPIPS}
        & \multicolumn{1}{c}{Time / sample} \\
        & \multicolumn{1}{c}{(dB) $\uparrow$}
        & \multicolumn{1}{c}{(dB) $\uparrow$}
        & \multicolumn{1}{c}{$\downarrow$}
        & \multicolumn{1}{c}{(s) $\downarrow$} \\
        \midrule
         2 & 30.303 & 16.829 & 0.1610 & 0.1813 \\
         5 & 30.121 & 16.621 & 0.1716 & 0.2897 \\
        10 & 30.519 & 16.599 & 0.1751 & 0.4579 \\
        20 & 30.098 & 16.579 & 0.1757 & 0.7840 \\
        50 & 30.193 & 16.534 & 0.1689 & 1.7994 \\
        \bottomrule
    \end{tabular}
\end{table}

\begin{table}[t]
    \centering
    \scriptsize   
    \setlength{\tabcolsep}{3.5pt}  
    \caption{ODE starting time $t^\star$ and end time $t_{\text{end}}$ for different SNR values in the AWGN channel on DIV2K dataset.}
    \begin{tabular}{@{}c|ccccccc@{}}
        \toprule
        SNR (dB)      & 0   & 3   & 5   & 7   & 10  & 12  & 15 \\
        \midrule
        $t^\star$     & 0.463 & 0.328 & 0.261 & 0.207 & 0.147 & 0.116 & 0.082 \\
        $t_{\text{end}}$ & 0.000 & 0.000 & 0.000 & 0.000 & 0.000 & 0.000 & 0.000 \\
        \bottomrule
    \end{tabular}
    \label{tab:snr_tstar}
\end{table}

\subsubsection{Analysis of the Scheduler}
Due to the implementation of the ODE solver in our code, the time scale used here is reversed compared with the earlier description: \(t=0\) corresponds to a clean image and \(t=1\) to pure noise. Using the same DIV2K image under multiple AWGN noise levels, Table~\ref{tab:snr_tstar} shows a clear monotonic dependence of the landing time \(t^\star\) on the SNR: higher SNR consistently leads to larger \(t^\star\). That is, the ODE integration can start closer to the noise-dominated end of the trajectory. The systematic trend across varying noise levels shows the effectiveness of the proposed design, with the landing time $t^\star$ providing a principled link between wireless channel conditions and FM dynamics, thereby enabling adaptive and interpretable decoding.



\section{Conclusions}
\label{sec:conclusion}

We proposed an LTT generative decoder for wireless image transmission, which embeds the physical channel into a continuous-time probability flow. By constructing a smoothing path for AWGN channels and training a conditional velocity field with CFM, the channel output is interpreted as a landing point on the path and deterministically transported to the clean image by solving an ODE, without stochastic diffusion sampling. Through MMSE-based preprocessing, Rayleigh and MIMO channels are converted into equivalent AWGN channels with calibrated landing points. Thus, the flow trained for AWGN channels can be used for Rayleigh and MIMO channels. Experiments on various datasets across various channels demonstrate the effectiveness of the proposed LTT decoder. 

\printbibliography

@article{lipman2022flow,
  title   = {Flow Matching for Generative Modeling},
  author  = {Lipman, Yaron and Chen, Ricky T. Q. and Ben-Hamu, Heli and Nickel, Maximilian and Le, Matt},
  journal = {arXiv preprint arXiv:2210.02747},
  year    = {2022},
  month   = oct
}

@article{lecun1998gradient,
  title   = {Gradient-Based Learning Applied to Document Recognition},
  author  = {LeCun, Yann and Bottou, L{\'e}on and Bengio, Yoshua and Haffner, Patrick},
  journal = {Proc. IEEE},
  volume  = {86},
  number  = {11},
  pages   = {2278--2324},
  year    = {1998},
  month   = nov,
  doi     = {10.1109/5.726791}
}

@article{xiao2017fashionmnist,
  title   = {Fashion-{MNIST}: A Novel Image Dataset for Benchmarking Machine Learning Algorithms},
  author  = {Xiao, Han and Rasul, Kashif and Vollgraf, Roland},
  journal = {arXiv preprint arXiv:1708.07747},
  year    = {2017},
  month   = aug
}

@inproceedings{agustsson2017ntire,
  title     = {{NTIRE} 2017 Challenge on Single Image Super-Resolution: Dataset and Study},
  author    = {Agustsson, Eirikur and Timofte, Radu},
  booktitle = {Proc. IEEE Conf. Comput. Vis. Pattern Recognit. Workshops (CVPRW)},
  pages     = {1122--1131},
  year      = {2017},
  month     = jul,
  doi       = {10.1109/CVPRW.2017.150}
}

@article{bourtsoulatze2019deep,
  title   = {Deep Joint Source-Channel Coding for Wireless Image Transmission},
  author  = {Bourtsoulatze, Eirina and Burth Kurka, David and G{\"u}nd{\"u}z, Deniz},
  journal = {IEEE Trans. Cogn. Commun. Netw.},
  volume  = {5},
  number  = {3},
  pages   = {567--579},
  year    = {2019},
  month   = sep,
  doi     = {10.1109/TCCN.2019.2919300}
}

@article{cddm,
  title   = {{CDDM}: Channel Denoising Diffusion Models for Wireless Semantic Communications},
  author  = {Wu, Tong and Chen, Zhiyong and He, Dazhi and Qian, Liang and Xu, Yin and Tao, Meixia and Zhang, Wenjun},
  journal = {IEEE Trans. Wireless Commun.},
  volume  = {23},
  number  = {9},
  pages   = {11168--11183},
  year    = {2024},
  month   = sep,
  doi     = {10.1109/TWC.2024.3379244}
}

@inproceedings{wang2003multiscale,
  title     = {Multi-scale Structural Similarity for Image Quality Assessment},
  author    = {Wang, Zhou and Simoncelli, Eero P. and Bovik, Alan C.},
  booktitle = {Proc. IEEE Asilomar Conf. Signals, Syst. Comput.},
  pages     = {1398--1402},
  year      = {2003},
  month     = nov,
  doi       = {10.1109/ACSSC.2003.1292216}
}

@inproceedings{zhang2018unreasonable,
  title     = {The Unreasonable Effectiveness of Deep Features as a Perceptual Metric},
  author    = {Zhang, Richard and Isola, Phillip and Efros, Alexei A. and Shechtman, Eli and Wang, Oliver},
  booktitle = {Proc. IEEE/CVF Conf. Comput. Vis. Pattern Recognit. (CVPR)},
  pages     = {586--595},
  year      = {2018},
  month     = jun,
  doi       = {10.1109/CVPR.2018.00068}
}

@inproceedings{ho2020denoising,
  title     = {Denoising Diffusion Probabilistic Models},
  author    = {Ho, Jonathan and Jain, Ajay and Abbeel, Pieter},
  booktitle = {Adv. Neural Inf. Process. Syst. (NeurIPS)},
  volume    = {33},
  pages     = {6840--6851},
  year      = {2020},
  month     = dec
}

@inproceedings{song2021score,
  title     = {Score-Based Generative Modeling through Stochastic Differential Equations},
  author    = {Song, Yang and Sohl-Dickstein, Jascha and Kingma, Diederik P. and Kumar, Abhishek and Ermon, Stefano and Poole, Ben},
  booktitle = {Int. Conf. Learn. Represent. (ICLR)},
  year      = {2021},
  month     = may
}

@inproceedings{song2023consistency,
  title     = {Consistency Models},
  author    = {Song, Yang and Dhariwal, Prafulla and Chen, Mark and Sutskever, Ilya},
  booktitle = {Proc. 40th Int. Conf. Mach. Learn. (ICML)},
  volume    = {202},
  pages     = {32211--32252},
  year      = {2023},
  month     = jul
}

@article{bourtsoulatze2019deepjscc,
  title   = {Deep Joint Source-Channel Coding for Wireless Image Transmission},
  author  = {Bourtsoulatze, Eirina and Burth Kurka, David and G{\"u}nd{\"u}z, Deniz},
  journal = {IEEE Trans. Cogn. Commun. Netw.},
  volume  = {5},
  number  = {3},
  pages   = {567--579},
  year    = {2019},
  month   = sep,
  doi     = {10.1109/TCCN.2019.2919300}
}

@article{guo2024diffsemcom,
  title   = {Diffusion-Driven Semantic Communication for Generative Models with Bandwidth Constraints},
  author  = {Guo, Lei and Chen, Wei and Sun, Yuxuan and Ai, Bo and Pappas, Nikolaos and Quek, Tony Q. S.},
  journal = {arXiv preprint arXiv:2407.18468},
  year    = {2024},
  month   = jul,
  doi     = {10.48550/arXiv.2407.18468}
}

@article{pei2025ldmsemcom,
  title   = {Latent Diffusion Model-Enabled Low-Latency Semantic Communication in the Presence of Semantic Ambiguities and Wireless Channel Noises},
  author  = {Pei, Jianhua and Feng, Cheng and Wang, Ping and Tabassum, Hina and Shi, Dongyuan},
  journal = {IEEE Trans. Wireless Commun.},
  volume  = {24},
  number  = {5},
  pages   = {4055--4072},
  year    = {2025},
  month   = may,
  doi     = {10.1109/TWC.2025.3535714}
}

@article{zhang2025sgdjscc,
  title   = {Semantics-Guided Diffusion for Deep Joint Source-Channel Coding in Wireless Image Transmission},
  author  = {Zhang, Maojun and Wu, Haotian and Zhu, Guangxu and Jin, Richeng and Chen, Xiaoming and G{\"u}nd{\"u}z, Deniz},
  journal = {arXiv preprint arXiv:2501.01138},
  year    = {2025},
  month   = jan,
  doi     = {10.48550/arXiv.2501.01138}
}

@article{grassucci2023generative,
  title   = {Generative Semantic Communication: Diffusion Models Beyond Bit Recovery},
  author  = {Grassucci, Eleonora and Barbarossa, Sergio and Comminiello, Danilo},
  journal = {arXiv preprint arXiv:2306.04321},
  year    = {2023},
  month   = jun,
  doi     = {10.48550/arXiv.2306.04321}
}

@misc{ETSI_EN_302_755,
  organization = {{ETSI}},
  title        = {{Digital Video Broadcasting (DVB); Frame structure channel coding and modulation for a second generation digital terrestrial television broadcasting system (DVB-T2)}},
  howpublished = {ETSI EN 302 755 V1.3.1},
  year         = {2012},
  address      = {Sophia Antipolis, France}
}

@article{wu2024deepjsccmimo,
  title   = {Deep Joint Source-Channel Coding for Adaptive Image Transmission Over {MIMO} Channels},
  author  = {Wu, Haotian and Shao, Yulin and Bian, Chenghong and Mikolajczyk, Krystian and G{\"u}nd{\"u}z, Deniz},
  journal = {IEEE Trans. Wireless Commun.},
  volume  = {23},
  number  = {10},
  pages   = {15002--15017},
  year    = {2024},
  month   = oct,
  doi     = {10.1109/TWC.2024.3422794}
}

@article{fu2025computation,
  title   = {Computation-Resource-Efficient Task-Oriented Communications},
  author  = {Fu, Jingwen and Xiao, Ming and Ren, Chao and Skoglund, Mikael},
  journal = {IEEE Trans. Commun.},
  volume  = {73},
  number  = {11},
  pages   = {10631--10646},
  year    = {2025},
  month   = nov,
  doi     = {10.1109/TCOMM.2025.3587076}
}

@article{gunduz2024jsccsurvey,
  title   = {Joint Source--Channel Coding: Fundamentals and Recent Progress in Practical Designs},
  author  = {G{\"u}nd{\"u}z, Deniz and Wigger, Mich{\`e}le A. and Tung, Tze-Yang and Zhang, Ping and Xiao, Yong},
  journal = {Proc. IEEE},
  year    = {2024},
  month   = dec,
  doi     = {10.1109/JPROC.2024.3477331},
  note    = {early access}
}

@article{suto2025semanticimage,
  title   = {Semantic Communication for Image Transmission},
  author  = {Suto, Katsuya},
  journal = {IEICE ESS Fundam. Rev.},
  volume  = {19},
  number  = {2},
  pages   = {70--77},
  year    = {2025},
  month   = apr,
  doi     = {10.1587/essfr.19.2_70}
}

@article{luong2025diffusion,
  title   = {Diffusion Models for Future Networks and Communications: A Comprehensive Survey},
  author  = {Luong, Nguyen Cong and Hai, Nguyen Duc and Van Le, Duc and Nguyen, Huy T. and Vu, Thai-Hoc and Huynh-The, Thien and Zhang, Ruichen and Nguyen Duc Duy, Anh and Niyato, Dusit and Di Renzo, Marco and others},
  journal = {arXiv preprint arXiv:2508.01586},
  year    = {2025},
  month   = aug,
  doi     = {10.48550/arXiv.2508.01586}
}

@article{fan2025generative,
  title   = {Generative Diffusion Models for Wireless Networks: Fundamental, Architecture, and State-of-the-Art},
  author  = {Fan, Dayu and Meng, Rui and Xu, Xiaodong and Liu, Yiming and Nan, Guoshun and Feng, Chenyuan and Han, Shujun and Gao, Song and Xu, Bingxuan and Niyato, Dusit and others},
  journal = {arXiv preprint arXiv:2507.16733},
  year    = {2025},
  month   = jul,
  doi     = {10.48550/arXiv.2507.16733}
}

@inproceedings{ronneberger2015u,
  title     = {{U-Net}: Convolutional Networks for Biomedical Image Segmentation},
  author    = {Ronneberger, Olaf and Fischer, Philipp and Brox, Thomas},
  booktitle = {Int. Conf. Med. Image Comput. Comput.-Assist. Intervent. (MICCAI)},
  pages     = {234--241},
  year      = {2015},
  month     = oct,
  organization = {Springer},
  doi       = {10.1007/978-3-319-24574-4_28}
}

@article{tong2024improving,
  author  = {Alexander Tong and Kilian Fatras and Nikolay Malkin
             and Guillaume Huguet and Yanlei Zhang and
             Jarrid Rector-Brooks and Guy Wolf and Yoshua Bengio},
  title   = {Improving and Generalizing Flow-based Generative Models
             with Minibatch Optimal Transport},
  journal = {Trans. Mach. Learn. Res.},
  year    = {2024},
  month   = mar,
  pages   = {1--34},
  doi     = {10.48550/arXiv.2302.00482},
  note    = {Also available as arXiv:2302.00482}
}

@inproceedings{pu2025art,
  author    = {Yifan Pu and Yiming Zhao and Zhicong Tang and Ruihong Yin and
               Haoxing Ye and Yuhui Yuan and Dong Chen and Jianmin Bao and
               Sirui Zhang and Yanbin Wang and Lin Liang and Lijuan Wang and
               Ji Li and Xiu Li and Zhouhui Lian and Gao Huang and Baining Guo},
  title     = {ART: Anonymous Region Transformer for Variable Multi-Layer
               Transparent Image Generation},
  booktitle = {Proc. IEEE/CVF Conf. Comput. Vis. Pattern Recognit. (CVPR)},
  year      = {2025},
  month     = jun,
  pages     = {7952--7962}
}

@ARTICLE{986011,
  author={Skoglund, M. and Phamdo, N. and Alajaji, F.},
  journal={IEEE Trans. Inf. Theory}, 
  title={Design and performance of VQ-based hybrid digital-analog joint source-channel codes}, 
  year={2002},
  volume={48},
  number={3},
  pages={708-720},
  doi={10.1109/18.986011}}

\appendices

\section{Proof of the Continuity Equation}
\label{app:continuity}

Recall the conditional Gaussian density
\begin{equation}
  p_{t|1}(x\mid x_1)
  = \frac{1}{(2\pi\sigma^2(t))^{d/2}}
    \exp\!\Big(-\frac{\|x-\mu_t(x_1)\|^2}{2\sigma^2(t)}\Big),
\end{equation}
with $\sigma(t)>0$, $\dot\sigma(t)=\frac{\mathrm d}{\mathrm dt}\sigma(t)$ and
$\dot\mu_t(x_1)=\frac{\mathrm d}{\mathrm dt}\mu_t(x_1)$, and the conditional velocity field
\begin{equation}
  u_t(x\mid x_1)
  = \frac{\dot\sigma(t)}{\sigma(t)}\bigl(x-\mu_t(x_1)\bigr)
    + \dot\mu_t(x_1).
\end{equation}
We show that $(p_{t|1},u_t)$ satisfies the continuity equation
\begin{equation}
\partial_t p_{t|1}(x\mid x_1)
+ \nabla\!\cdot\!\big(p_{t|1}(\cdot\mid x_1)\,u_t(\cdot\mid x_1)\big)(x)
= 0.
\label{eq:CE-conditional}
\end{equation}

For brevity, writing $p_{t|1}=p_{t|1}(x\mid x_1)$, $\mu_t=\mu_t(x_1)$ and
$\sigma=\sigma(t)$, we have
\begin{equation}
  \log p_{t|1}
  = -\frac d2\log(2\pi\sigma^2)
    -\frac{\|x-\mu_t\|^2}{2\sigma^2},
\end{equation}
and thus
\begin{equation}
  \partial_t \log p_{t|1}
  = -\frac{d\,\dot\sigma}{\sigma}
    + \frac{(x-\mu_t)\!\cdot\!\dot\mu_t}{\sigma^2}
    + \frac{\|x-\mu_t\|^2\,\dot\sigma}{\sigma^3},
\end{equation}
and
\begin{equation}
  \nabla \log p_{t|1}
  = -\frac{x-\mu_t}{\sigma^2},
\end{equation}
\begin{equation}
  \nabla p_{t|1}
  = p_{t|1}\,\nabla\log p_{t|1}
  = -\frac{x-\mu_t}{\sigma^2}\,p_{t|1}.    
\end{equation}

Therefore, we have
\begin{equation}
  \partial_t p_{t|1}
  = p_{t|1}\,\partial_t \log p_{t|1}
  = p_{t|1}\Big[
      -d\,\tfrac{\dot\sigma}{\sigma}
      + \tfrac{(x-\mu_t)\!\cdot\!\dot\mu_t}{\sigma^2}
      + \tfrac{\|x-\mu_t\|^2\dot\sigma}{\sigma^3}
    \Big].
\end{equation}

Since $\sigma(t)$ and $\mu_t(x_1)$ do not depend on $x$, we obtain
\begin{align}
\nabla\!\cdot\!\big(p_{t|1}u_t\big)
 &= \nabla\!\cdot\!\Big(
      p_{t|1}\,\tfrac{\dot\sigma}{\sigma}(x-\mu_t)
    \Big)
  + \nabla\!\cdot\!\big(p_{t|1}\dot\mu_t\big) \nonumber\\
 &= \tfrac{\dot\sigma}{\sigma}\Big[
      d\,p_{t|1}
      + (x-\mu_t)\!\cdot\!\nabla p_{t|1}
    \Big]
    + \dot\mu_t\!\cdot\!\nabla p_{t|1} \nonumber\\
 &= \tfrac{\dot\sigma}{\sigma}\Big[
      d
      - \tfrac{\|x-\mu_t\|^2}{\sigma^2}
    \Big]p_{t|1}
    - \tfrac{(x-\mu_t)\!\cdot\!\dot\mu_t}{\sigma^2}\,p_{t|1}.
\end{align}
Adding the two expressions yields
\begin{align}
\partial_t p_{t|1} + \nabla\!\cdot\!(p_{t|1}u_t)
&= p_{t|1}\Big[
   -d\,\tfrac{\dot\sigma}{\sigma}
   + \tfrac{(x-\mu_t)\!\cdot\!\dot\mu_t}{\sigma^2}
   + \tfrac{\|x-\mu_t\|^2\dot\sigma}{\sigma^3} \nonumber\\
&\quad\ \ 
   + \tfrac{\dot\sigma}{\sigma}\Big(d-\tfrac{\|x-\mu_t\|^2}{\sigma^2}\Big)
   - \tfrac{(x-\mu_t)\!\cdot\!\dot\mu_t}{\sigma^2}
  \Big] = 0,
\end{align}
which proves~\eqref{eq:CE-conditional}.
\hfill$\square$

\section{Proof of Proposition~\ref{prop:scalar-gaussian}}
\label{app:proof-prop1}

Under the settings, Gaussian path satisfies
\(
X_t \sim \mathcal N(0,s^2(t))
\)
for all \(t\in[0,1]\).
The continuity equation
\begin{equation}
    \partial_t p_t(x) + \partial_x\big(p_t(x)\,v_t(x)\big) = 0
\end{equation}
is satisfied by the velocity field \(v_t(x) = \frac{\dot s(t)}{s(t)}x\). Consequently, the ODE
\begin{equation}
    \frac{\mathrm{d}}{\mathrm{d}t} X_t = \frac{\dot s(t)}{s(t)} X_t
\end{equation}
has the solution
\begin{equation}
    X_t = X_{t^\star} \frac{s(t)}{s(t^\star)},
    \qquad t\in[t^\star,1].
\end{equation}
Evaluating at \(t=1\) and using \(s(1)=\sigma_x\) and \(s(t^\star)=\sqrt{\sigma_x^2+\sigma_{\mathrm{ch}}^2}\) yields
\begin{equation}
    \widehat X_1^{\mathrm{LTT}}
    = X_1(t=1)
    = \frac{\sigma_x}{\sqrt{\sigma_x^2 + \sigma_{\mathrm{ch}}^2}}\,Y
    = a_{\mathrm{LTT}} Y,
\end{equation}
where
\(
a_{\mathrm{LTT}} = \frac{\sigma_x}{\sqrt{\sigma_x^2 + \sigma_{\mathrm{ch}}^2}}.
\)
The linear MMSE estimator in the scalar Gaussian model is well known to be
\begin{equation}
    a_{\mathrm{MMSE}} = \frac{\mathrm{Cov}(X_1,Y)}{\mathrm{Var}(Y)}
    = \frac{\sigma_x^2}{\sigma_x^2 + \sigma_{\mathrm{ch}}^2},
\end{equation}
with MSE
\begin{equation}
    \mathrm{MSE}_{\mathrm{MMSE}}
    = \sigma_x^2 - \frac{\sigma_x^4}{\sigma_x^2 + \sigma_{\mathrm{ch}}^2}
    = \frac{\sigma_x^2\,\sigma_{\mathrm{ch}}^2}{\sigma_x^2 + \sigma_{\mathrm{ch}}^2}.
\end{equation}
For a generic linear estimator \(\widehat X_1 = a Y\), the MSE can be written as
\begin{equation}
    \mathrm{MSE}(a)
    = \sigma_x^2 - 2 a \sigma_x^2 + a^2(\sigma_x^2 + \sigma_{\mathrm{ch}}^2).
\end{equation}
A standard quadratic expansion shows that
\begin{equation}
    \mathrm{MSE}(a)
    = \mathrm{MSE}_{\mathrm{MMSE}}
      + (\sigma_x^2 + \sigma_{\mathrm{ch}}^2)(a - a_{\mathrm{MMSE}})^2.
\end{equation}
Substituting \(a = a_{\mathrm{LTT}}\) gives the expression for
\(
\mathrm{MSE}_{\mathrm{LTT}} \triangleq \mathrm{MSE}(a_{\mathrm{LTT}})
\).
A Taylor expansion of \(a_{\mathrm{LTT}}\) and \(a_{\mathrm{MMSE}}\) around \(\sigma_{\mathrm{ch}} = 0\) yields
\begin{equation}
    (\sigma_x^2 + \sigma_{\mathrm{ch}}^2)\big(a_{\mathrm{LTT}} - a_{\mathrm{MMSE}}\big)^2
    = \frac{\sigma_{\mathrm{ch}}^4}{4\,\sigma_x^2}
      + o\!\big(\sigma_{\mathrm{ch}}^4\big),
\end{equation}
which implies
\(
\mathrm{MSE}_{\mathrm{LTT}} - \mathrm{MSE}_{\mathrm{MMSE}}
    = o(\sigma_{\mathrm{ch}}^4)
\)
and completes the proof.
\hfill$\square$

\section{Proof of Proposition~\ref{prop:euler-complexity}}
\label{app:proof-prop2}

Under Assumption~\ref{assump:lipschitz-bounded},
let $h = T/N$ and $t_k = t^\star + kh$, $k=0,\dots,N$, and define Euler iterates $x_{k+1} = x_k + h f(x_k,t_k)$ with $x_0 = y$ and the global error $e_k = x_{\mathrm{cont}}(t_k;y) - x_k$. Using the integral form of the exact solution and subtracting Euler update yields
\begin{equation}
    e_{k+1}
    = e_k + \int_{t_k}^{t_{k+1}}\!\big[
            f\big(x_{\mathrm{cont}}(s;y),s\big)
            - f(x_k,t_k)
         \big]\mathrm ds.
\end{equation}
By the Lipschitz property and boundedness of $f$, one obtains
\begin{equation}
    \|e_{k+1}\| \le (1+Lh)\|e_k\| + LB h^2.
\end{equation}
Iterating this recursion with $e_0=0$ and applying Grönwall's inequality gives
\(
    \max_{0\le k\le N} \|e_k\|\le C h
\)
for some $C>0$ depending only on $L,B,T$. Since $h=T/N$, the stated bound follows.
\hfill$\square$

\section{Proof of Proposition~\ref{prop:convergence-rate}}
\label{app:proof-prop3}
Under Assumption~\ref{assump:lipschitz-bounded} and Proposition~\ref{prop:euler-complexity}, using
\begin{equation}
\begin{aligned}
    & \|X_1 - \widehat X_1^{(N)}(Y)\|^2 \\
    &= \|X_1 - \widehat X_1^{\mathrm{cont}}(Y) + \widehat X_1^{\mathrm{cont}}(Y) - \widehat X_1^{(N)}(Y)\|^2 \\
    &\le \big(\|X_1 - \widehat X_1^{\mathrm{cont}}(Y)\| + \|\widehat X_1^{\mathrm{cont}}(Y) - \widehat X_1^{(N)}(Y)\|\big)^2
\end{aligned}
\end{equation}
and taking expectations, Proposition~2 together with Cauchy--Schwarz yields
\begin{equation}
\begin{aligned}
    \mathrm{MSE}_{N}
    &\le \mathrm{MSE}_{\mathrm{cont}}
       + \frac{2C}{N}\sqrt{\mathrm{MSE}_{\mathrm{cont}}}
       + \frac{C^2}{N^2},
\end{aligned}
\end{equation}
where we used $\|\widehat X_1^{\mathrm{cont}}(Y) - \widehat X_1^{(N)}(Y)\|\le C/N$ almost surely. In particular,
\begin{equation}
    \mathrm{MSE}_{N} - \mathrm{MSE}_{\mathrm{cont}} = \mathcal O\!\big(\tfrac{1}{N}\big),
\end{equation}
so the distortion of the discretized decoder converges to that of the continuous-time ODE decoder at rate $1/N$ as the number of ODE steps increases.
\hfill\(\square\)

\end{document}